\documentclass[12pt]{article}
\usepackage{arxiv}
\usepackage[numbers]{natbib}
\usepackage{hyperref}  
\usepackage[table]{xcolor}  
\usepackage{pgf} 

\RequirePackage{graphicx}
\usepackage{xcolor}
\usepackage{natbib}
\usepackage{subfig}
\usepackage{caption}
\usepackage{subcaption}

\usepackage{listings}    
\usepackage{amsmath,amsfonts,bm}

\def\eqref#1{equation~\ref{#1}}

\def\1{\bm{1}}

\DeclareMathAlphabet{\mathsfit}{\encodingdefault}{\sfdefault}{m}{sl}
\SetMathAlphabet{\mathsfit}{bold}{\encodingdefault}{\sfdefault}{bx}{n}

\newcommand{\ours}{MALADE\xspace}
\newcommand{\oursfull}{(\textbf{M}ultiple \textbf{A}gents powered by \textbf{L}LMs for \textbf{ADE} Extraction)\xspace}

\newcommand{\Task}{\texttt{Task}}
\newcommand{\task}{\texttt{task}}

\newcommand{\Agent}{\texttt{Agent}}
\newcommand{\stringtostring}{\texttt{string $\rightarrow$ string}}
\newcommand{\llmresponse}{\texttt{llm\_response}~}
\newcommand{\userresponse}{\texttt{user\_response}~}
\newcommand{\agentresponse}{\texttt{agent\_response}~}

\newcommand{\drugagent}{\texttt{DrugFinder}\xspace}
\newcommand{\interactionagent}{\texttt{DrugAgent}\xspace}
\newcommand{\fdahandler}{\texttt{FDAHandler}\xspace}
\newcommand{\labelagent}{\texttt{CategoryAgent}\xspace}

\newcommand{\ben}{\begin{enumerate}}
\newcommand{\een}{\end{enumerate}}
\newcommand{\beq}{\begin{equation}}
\newcommand{\eeq}{\end{equation}}
\newcommand{\beqa}{\begin{eqnarray}}
\newcommand{\eeqa}{\end{eqnarray}}
\newcommand{\bit}{\begin{itemize}}
\newcommand{\eit}{\end{itemize}}
\newcommand{\btab}{\begin{tabular}}
\newcommand{\etab}{\end{tabular}}

\newcommand{\noprint}[1]{}

\def \ie {{\em i.e.},~}

\def \eg {{\em e.g.},~}

\def \etal {{\em et al.}}

\newcommand{\code}[1]{\texttt{#1}}
\def \cross {\,$\times$\,}
\def \to {\,$\rightarrow$\,}

\usepackage{adjustbox}
\usepackage{longtable}
\usepackage{multirow}
\usepackage{subcaption}
\usepackage{enumitem}
\usepackage{fancyvrb}
\usepackage{changepage}

\usepackage{booktabs} %
\usepackage{xspace}

\usepackage{calc} %

\title{\ours: Orchestration of LLM-powered Agents \\ with Retrieval Augmented Generation for Pharmacovigilance}
\author{
Jihye Choi\thanks{Equal contribution, listed alphabetically by last name} \hspace{1mm}$^1$,   
Nils Palumbo\footnotemark[1] \hspace{1mm}$^1$,
Prasad Chalasani$^2$, 
Matthew M. Engelhard$^3$, \\
\bf Somesh Jha$^{1,2}$,
\bf Anivarya Kumar$^3$, 
\bf David Page$^3$
\\
$^1$University of Wisconsin-Madison, $^2$Langroid, $^3$Duke University}

  

\begin{document}

\maketitle

\begin{abstract}

In the era of Large Language Models (LLMs), given their remarkable text understanding and generation abilities, there is an unprecedented opportunity
to develop new, LLM-based methods for
trustworthy medical knowledge synthesis, extraction and summarization.
This paper focuses on the problem of Pharmacovigilance (PhV), where the significance and challenges lie in
identifying Adverse Drug Events (ADEs) from diverse text sources, such as medical literature, clinical notes, and drug labels.
Unfortunately, this task is hindered by factors including variations in the terminologies of drugs and outcomes, and ADE descriptions often being buried in large amounts of narrative text.
We present \ours, the first effective collaborative multi-agent system powered by LLM with Retrieval Augmented Generation for ADE extraction from drug label data.
This technique involves augmenting a query to an LLM with relevant information extracted from text resources, and instructing the LLM to compose a response consistent with the augmented data.
\ours is a general LLM-agnostic architecture, and its unique capabilities are: {(1) leveraging a variety of external sources, such as medical literature, drug labels, and FDA tools (\eg OpenFDA drug information API), (2) extracting drug-outcome association in a structured format along with the strength of the association, and (3) providing explanations for established associations.} 
Instantiated with GPT-4 Turbo {or GPT-4o,} and FDA drug label data, \ours demonstrates its efficacy with an Area Under ROC Curve of {0.90} against the OMOP Ground Truth table of ADEs. 
Our implementation leverages the
{\href{https://github.com/langroid/langroid}{Langroid}} multi-agent LLM framework
and can be found at
{\url{https://github.com/jihyechoi77/malade}}.

\end{abstract}

\vspace{.2in}
\section{Introduction}

\begin{figure}[ht]
    \centering
    \includegraphics[width=\linewidth]{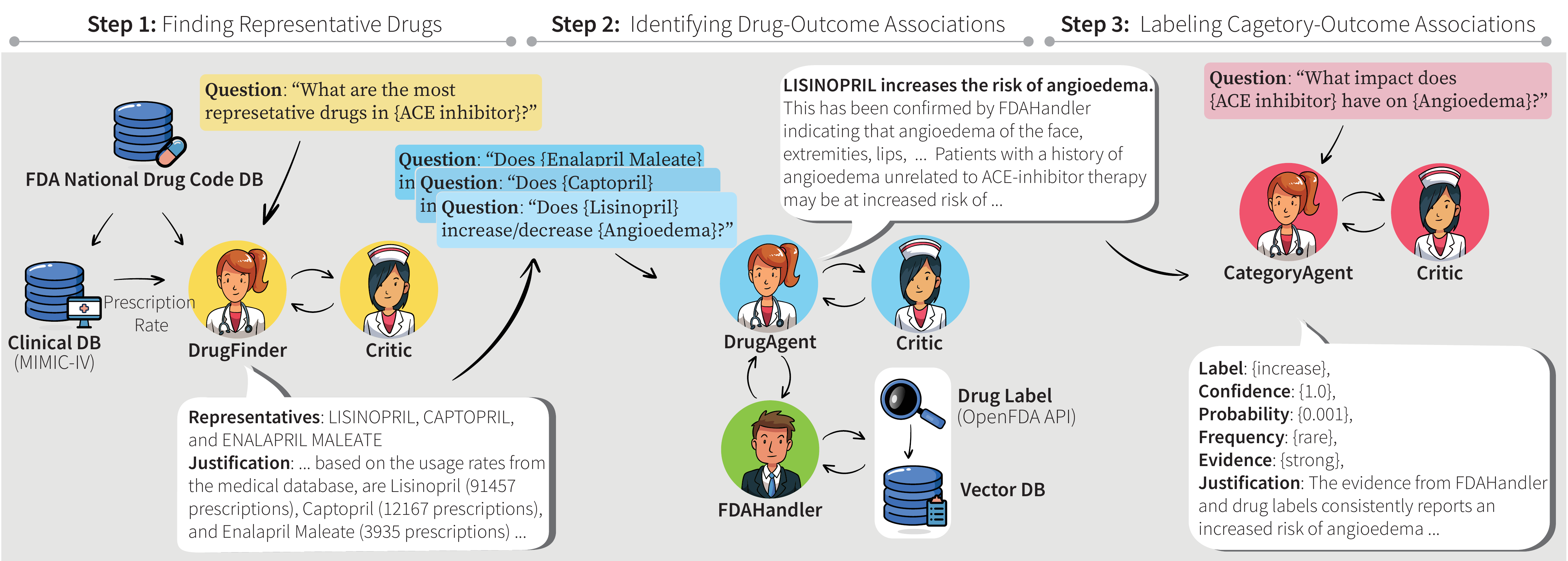}
    \caption{{\bf Real-world demonstration of our proposed multi-agent orchestration system, \ours.}
        Handling the user query, ``Are ACE Inhibitors associated with Angioedema?'',
        involves a sequence of subtasks performed by three Agents:
        \drugagent, \interactionagent, \labelagent (each instantiated with GPT-4 Turbo {or GPT-4o}).
        Each Agent generates a response and justification, which are validated by a corresponding Critic agent,
        whose feedback is used by the Agent to revise its response.}
    \label{fig:ours_intro}
\end{figure}

Pharmacovigilance (PhV) is the science of identification and prevention of adverse drug events (ADEs) caused by pharmaceutical products after they are introduced to the market.
PhV is of enormous importance to both the pharmaceutical industry and public health, as it
aims to safeguard the well-being of patients by detecting new safety concerns and intervening
when necessary.
A central problem in PhV is {\bf ADE Extraction}:
given a {\em drug category\/} $C$ and an {\em adverse event\/} $E$, determine whether
(and how strongly) $C$ is associated with $E$.
This task demands the analysis of a vast corpus of textual data sources from a variety of
sources, such as patient medical records, clinical notes, social media,
spontaneous reporting systems, drug labels, medical literature,
and clinical trial reports.
Besides the sheer volume of text from these sources, ADE extraction is further complicated by
variability in the names of drugs and outcomes, and
the fact that ADE descriptions are often buried in large amounts of narrative text~\citep{10.3389/fbinf.2023.1328613}.

Traditionally, various classical natural language processing (NLP)
and deep learning techniques have been used to address this problem~\citep{natarajan2017markov, mower2018learning, tiftikci2019machine, bayer2021ade}.
Compared to classical NLP methods, today's best Large Language Models (LLMs)
(and even weaker open-source/local LLMs~\citep{touvron2023llama,jiang2023mistral})
{exhibit a significant advancement}
in text understanding and generation capabilities,
and there is a great opportunity to use these models
to not only improve existing ADE extraction methods,
but also consider data sources that were previously not feasible to use.
Recent attempts to apply LLMs to ADE Extraction only leverage off-the-shelf ChatGPT~\citep{Wang+Ding+Luo2023},
with limited performance and inconsistent reasoning for their extraction rationales~\citep{sun-etal-2024-leveraging}.
These limitations stem primarily from two factors: (a) accurate ADE Extraction requires access to
specific data sources which LLMs may not have ``seen" during their pre-training,  
hence relying on an LLM's ``built-in" knowledge yields inaccurate results, and
(b) LLMs, being probabilistic next-token predictors, may produce
incorrect or unreliable results when used naively without carefully breaking down the task into simpler sub-tasks, or without mechanisms to validate
and correct their responses.

In this paper, we introduce \ours\footnote{Pronounced like the French word \textit{malade} meaning ``sick'' or ``ill.''}\oursfull,
the first effective multi-agent Retrieval-Augmented Generation
(RAG) system for ADE Extraction.
Our approach leverages two key techniques to address the above two limitations respectively:
(a) RAG, equipping an LLM with up-to-date knowledge by augmenting an input
query with relevant portions of text data, and prompting the LLM to generate responses
consistent with the augmented information~\citep{lewis2020retrieval};
and (b) strategic orchestration of multiple LLM-based agents,
each responsible for a relatively smaller sub-task of the overall ADE Extraction task~\citep{wu2023autogen}. 
Specifically, our system has agents for these sub-tasks (see Figure~\ref{fig:ours_intro}):
(1) identifying representative drugs for each drug category from
a medical database (\eg MIMIC-IV),
(2) gathering information on side effects of those drugs from external
text knowledge bases (\eg FDA drug label database), and finally,
(3) composing final answers summarizing the effect of the drug category on an adverse event.
Each agent is assigned a specific sub-task and collaborates with others to accomplish the
the ultimate goal of ADE identification.
Furthermore, we enhance the reliability of our multi-agent system even further by pairing each agent with a critic agent,
whose role is to verify the behaviors and responses of its counterpart.

The system, though applied here for ADE extraction specifically, illustrates how a Multi-Agent approach can be used to generate trustworthy, evidence-based summaries and confidence scores in response to challenging medical questions requiring synthesis of evidence from multiple sources of clinical knowledge and data. As such, \ours may be viewed as a case study illustrating an approach that could later be applied to other problems in PhV, including identification of possible drug-to-drug interactions, as well as clinical problems outside of PhV, such as identifying known symptoms of a condition of interest documented in clinical notes.

In summary, our paper makes the following contributions.
{
\paragraph{Precise Evaluation.}
In contrast to simpler systems that only produce a binary label indicating whether or not a drug category $C$ is associated with an adverse event $E$, 
our method produces distinct scores, including a confidence score that indicates how confident an LLM is about its label assignment.  
These scores permit a rigorous quantitative evaluation against the well-established 
Observational Medical Outcomes Partnership (OMOP) Ground Truth table of ADEs associated with common drug classes~\citep{omop}.
We achieve an Area Under the ROC Curve (AUC) of approximately 0.85 with GPT-4 Turbo, and 0.90 with GPT-4o (Section~\ref{sec:experimental}). 
To the best of our knowledge, this is the best performance among the baselines, even though the direct comparison may be limited~\footnote{{Because none of the original clinical data-based analyses reached this high of accuracy, followup investigations have since argued that roughly this level is the best achievable by any method based on any sources for the OMOP task.
In 2016, Gruber \etal~\cite{gruber2016design} argued there were reproducible errors that could be blamed on the OMOP 2010 ground truth itself that could place a ceiling on the AUC achievable, and Hauben \etal~\cite{hauben2016evidence} more specifically argued that on the negative-labeled drug event pairs the error in the ground truth should be estimated at 17\%.
There may be disagreement on varying strengths of different literature evidence, but
if their estimate is exactly right, it could place a ceiling as low as 0.83 on the AUC achievable.}}.
}

\paragraph{Grounded generation of responses and justifications.}
The design of \ours offers key features essential for
high-stakes applications like ADE identification:
(1) A structured format for drug-to-outcome associations, including scores
indicating the strength of the association and rarity of the adverse event;
this is important to ensure robust downstream processing of the extracted associations.
(2) Justifications for the extracted drug-outcome associations,
allowing human experts to understand and validate the associations.
This is possible due to the RAG component of the \ours architecture,
which allows leveraging various external sources such as medical literature, drug labels,
FDA tools (\eg OpenFDA drug information API), as well as common
clinical data sources such as OMOP or PCORI, and even specific EHR systems where available.
(3) Observability, \ie complete, detailed logs of inter-agent dialogs and intermediate steps;
these are essential for debugging and auditing the system's behavior.
See Figure~\ref{fig:ours_intro} for a real-world demonstration of \ours.

\paragraph{{Generalizable Insights about Machine Learning in the Context of Healthcare.}}
Our proposed multi-agent architecture is agnostic to LLMs and data sources and is based
on design primitives intended to be universal building blocks for the orchestration of
multiple LLM-based agents (Section~\ref{sec:preliminaries}). 
Hence, although \ours is instantiated specifically for ADE identification, 
our design methodology provides a generalizable blueprint for the
effective construction of multi-agent systems
for trustworthy medical knowledge synthesis and summarization with wide-ranging medical applications.

\section{Related Work}

The advent of highly-capable Large Language Models (LLMs) has sparked significant
interest in applying these models to medical tasks, including diagnostics~\citep{singhal2023large}, medical question-answering~\citep{
    singhal2023towards, nori2023capabilities}, and medical evidence summarization~\citep{tang2023evaluating}.
An important application area is pharmacovigilance, the science of identifying and preventing adverse drug events
(ADEs) caused by pharmaceutical products after they are introduced to the market.
The specific problem of ADE Extraction, namely, identifying whether a specific drug (or category) is associated with a
specific adverse event, is a challenging task due to variations in drug and outcome terminologies, the presence of ADE
descriptions in large amounts of narrative text, and the disparate sources of such text data, which
can include patient medical records, clinical notes, drug labels, medical literature, clinical trials,
message boards, social media. Prior research in this field, notably works drawing on large-scale research initiatives including Sentinel~\citep{platt2009new}, OMOP~\citep{ryan2013defining}, and OHDSI~\citep{stang2010advancing}, has focused on developing new methods for causal discovery from purely observational data.
Huang \etal~\cite{healthcare10040618} investigate the use of social forums for constructing predictive models of ADEs, focusing on the performance of different data processing techniques and BERT architectures.
von Csefalvay~\cite{von2024daedra} introduces a novel LLM, DAEDRA, for detecting regulatory-relevant outcomes from passive pharmacovigilance reports.
Sorbello \etal~\cite{sorbello2023} use LLMs like GPT to improve the capture of opioid drug and adverse event mentions from electronic health records.
Finally, Sun \etal~\cite{sun-etal-2024-leveraging} investigate the performance of ChatGPT for extracting adverse events from medical text sources.

These early applications of LLMs to ADE Extraction are limited in at least one of two ways:
(a) they either use only the bare LLM (such as ChatGPT, or its API)
without access to any external APIs, tools, or knowledge bases~\citep{Wang+Ding+Luo2023}. ADE extraction using only the LLM's ``built-in'' knowledge
(\ie text it was exposed to during pre-training) is likely to be inaccurate and incomplete,
since adverse events may be discovered in any new studies or reports;
(b) all prior works use a single LLM (even when augmented with external data/tools)
without any collaboration or feedback from other LLMs. Since LLMs are after all
probabilistic next-token prediction models, there is no guarantee that the generated text is accurate or complete.
The only way to improve the reliability of an LLM's responses in this scenario is to either
resort to elaborate prompting techniques~\citep{wei2022chain, yao2024tree},
or have a human {(or an LLM~\cite{self-refine})} in the loop to verify the generated text and iteratively refine the prompts until a satisfactory response is obtained.

To address these limitations, three paradigms have emerged in LLM practitioners' toolboxes.
The first limitation is addressed by two techniques: {\em Retrieval Augmented Generation} (RAG)
and {\em tool-use}.
RAG addresses the knowledge limitations of LLMs by augmenting the input prompt or query
with relevant information retrieved from external knowledge bases (using similarity based on
vector embeddings, keywords, or a combination of both), and instructing the LLM to respond
to the original query in a way that is consistent with the augmented data, and also to provide
a justification for its response by citing the relevant external data
~\citep{lewis2020retrieval}.
Thus the RAG approach not only alleviates the limitations of relying only on an LLM's pre-trained knowledge,
but also provides evidence-citation ability, which is crucial to engender trust in the LLM's
responses, especially in high-stakes applications like medical decision-making.
This approach has shown promise in enhancing LLM performance in biomedicine, particularly in literature information-seeking and clinical decision-making~\citep{frisoni2022bioreader, jin2023retrieve, wang2023augmenting, Almanac2024}.
The second technique, tool-use, involves instructing the LLM to produce {\em structured\/} text
(typically JSON) which can then be easily parsed by downstream code to perform a variety of actions,
including web-search, querying APIs for information, querying databases, and performing computations~\citep{ruan2023tptu, li-etal-2023-api}.

The emergence of multi-agent systems  addresses the second limitation (of using single LLMs) --
this approach aims to harness the collective capabilities of multiple LLMs~\citep{xi2023rise, hong2024metagpt}.
Such systems introduce cooperative learning and feedback mechanisms between LLM-based agents, which simulate human-like communication, consultation and debate processes, enabling them to tackle even more complex tasks than
a single-agent with RAG.
In medical reasoning tasks, for instance, multi-agent collaboration can mirror hospital
consultation mechanisms~\citep{tang2023medagents}.
Our work extends this trajectory of research; to the best of our knowledge, our system \ours is first effective multi-agent orchestration system with RAG and tool-use, tailored for a specific task in pharmacovigilance, namely ADE Extraction.
In our approach, LLM-based agents collaborate, leveraging their collective expertise and the
latest medical knowledge. This approach aims to improve the analysis of ADEs, offering a more robust and reliable
system for pharmacovigilance.

\section{Preliminaries on LLM-based Agents}
\label{sec:preliminaries}

While today's LLMs exhibit impressive capabilities, they remain constrained by technical and practical limitations such as brittleness, non-determinism, limited context window, inference costs, and latency~\citep{liang2023holistic},
with the implication that one cannot simply give high-level instructions to an LLM and expect it to accomplish a complex task.
Consequently, to best harness the capabilities of LLMs as components of a complex application, it is necessary to decompose the task into smaller sub-tasks and manage multiple LLM conversations, each with its own set of specifically-defined instructions, state, and data sources. This leads naturally to the notion of an {\em agent\/} as an LLM-powered entity responsible for a well-defined small sub-task.
In Section~\ref{sec:background-agent-oriented}, we introduce the key abstractions and
components needed for agent-oriented programming, and
Section~\ref{sec:background-multi-agent} describes multi-agent orchestration.
{Our implementation leverages the open-source multi-agent LLM framework
\href{https://github/langroid/langroid}{Langroid}~\citep{langroid2023},
which supports these abstractions and mechanisms}.

\subsection{Agent-oriented Programming}
\label{sec:background-agent-oriented}

\paragraph{Agent, as an intelligent message transformer.} A natural and convenient abstraction in designing a complex
LLM-powered system is the notion of an \textit{agent} that is instructed to be responsible for a specific aspect of the overall task. In terms of code, an \texttt{Agent} is essentially a class representing an intelligent entity that can respond to \textit{messages}, \ie an agent is simply a \textit{message transformer}.
An agent typically encapsulates an (interface to an) LLM, and may also be equipped with so-called \textit{tools} (as described below) and
\textit{external documents/data} (\eg a vector database, as described below).
Much like a team of humans, agents interact by exchanging messages, in a manner reminiscent of the \textit{actor framework} in programming languages~\citep{hewitt_actor_2010}.
An \textit{orchestration mechanism} is needed to manage the flow of messages between agents, to ensure that progress is made towards completion of the task, and to handle the inevitable cases where an agent deviates from instructions.
In this work we adopt this \textit{multi-agent programming} paradigm, where agents are first-class citizens, acting as message transformers, and communicate by exchanging messages.

To build useful applications with LLMs, we need to endow them with the ability to
trigger actions (such as API calls, computations, database queries, etc) and
access external documents. \textit{Tools} and \textit{Retrieval Augmented Generation} (RAG)
provide these capabilities, described next.

\paragraph{Tools, also known as functions or plugins.}
An LLM is essentially a text transformer; \ie in response to some input text (known as a \textit{prompt}), it produces a response. Free-form text responses are ideal when we want to generate a description, answer, or summary for human consumption, or even a question for another agent to answer.
However, in some cases, we would like the responses to trigger external \textit{actions}, such as an API call, code execution, or a database query. In such cases, we would instruct the LLM to produce a \textit{structured} output, typically in JSON format, with various pre-specified fields, such as code, an SQL query, parameters of an API call, and so on. These structured responses have come to be known as {\em tools}, and the LLM is said to \textit{use} a tool when it produces a structured response corresponding to a specific tool. To elicit a tool response from an LLM, it needs to be instructed on the expected tool format and the conditions under which it should use the tool.
To actually use a tool emitted by an LLM, a \textit{tool handler} method must be defined as well.
The tool handler for a given tool is triggered when it is recognized in the LLM's response.
See Appendix~\ref{app:tool-example} for a description of the LLM's interaction with a database.

Starting with the view of an LLM as a text transformer, it turns out that one can express the notion
of an agent, a tool, and other related concepts in terms of different function signatures,
as shown in Table~\ref{tab:llm-agent-typesig} in Appendix~\ref{app:llm-to-agent}.

\paragraph{Retrieval Augmented Generation (RAG).}
Using an LLM in isolation has two major constraints: (a) the responses are confined to the knowledge from its pre-training, hence cannot answer questions specific to private/enterprise documents, or up-to-date information past its training cutoff date; and (b) there is no way to verify the validity of the generated answers.
RAG is the most popular technique to address both limitations by making LLMs generate responses based on specific documents or data and justify the answer by presenting source citations~\citep{lewis2020retrieval}.
The basic idea of RAG is as follows: when a query $Q$ is made to an LLM-agent, a set of $k$ documents (or portions thereof)
$D = \{d_1, d_2, \ldots, d_k\}$ most ``relevant'' to the query are \textit{retrieved} from a document-store,
and the original query $Q$ is \textit{augmented} with $D$ to a new prompt of the form,
``Given the passages below: [$d_1$, $d_2$, \ldots, $d_k$], answer this question: $Q$ based ONLY on these passages,
and indicate which passages support your answer''.
See Appendix~\ref{app:rag} for more details on RAG.

\subsection{Multi-Agent Orchestration}
\label{sec:background-multi-agent}
As mentioned above, when building an LLM-based multi-agent system, an orchestration mechanism is critical to manage the flow of messages between agents, to ensure task progress, and handle deviations from instructions.
In this work, we leverage a simple yet versatile orchestration mechanism that seamlessly handles user interaction, tool handling, and sub-task delegation.
\begin{figure}[t]
    \centering
    \includegraphics[width=.85\textwidth]{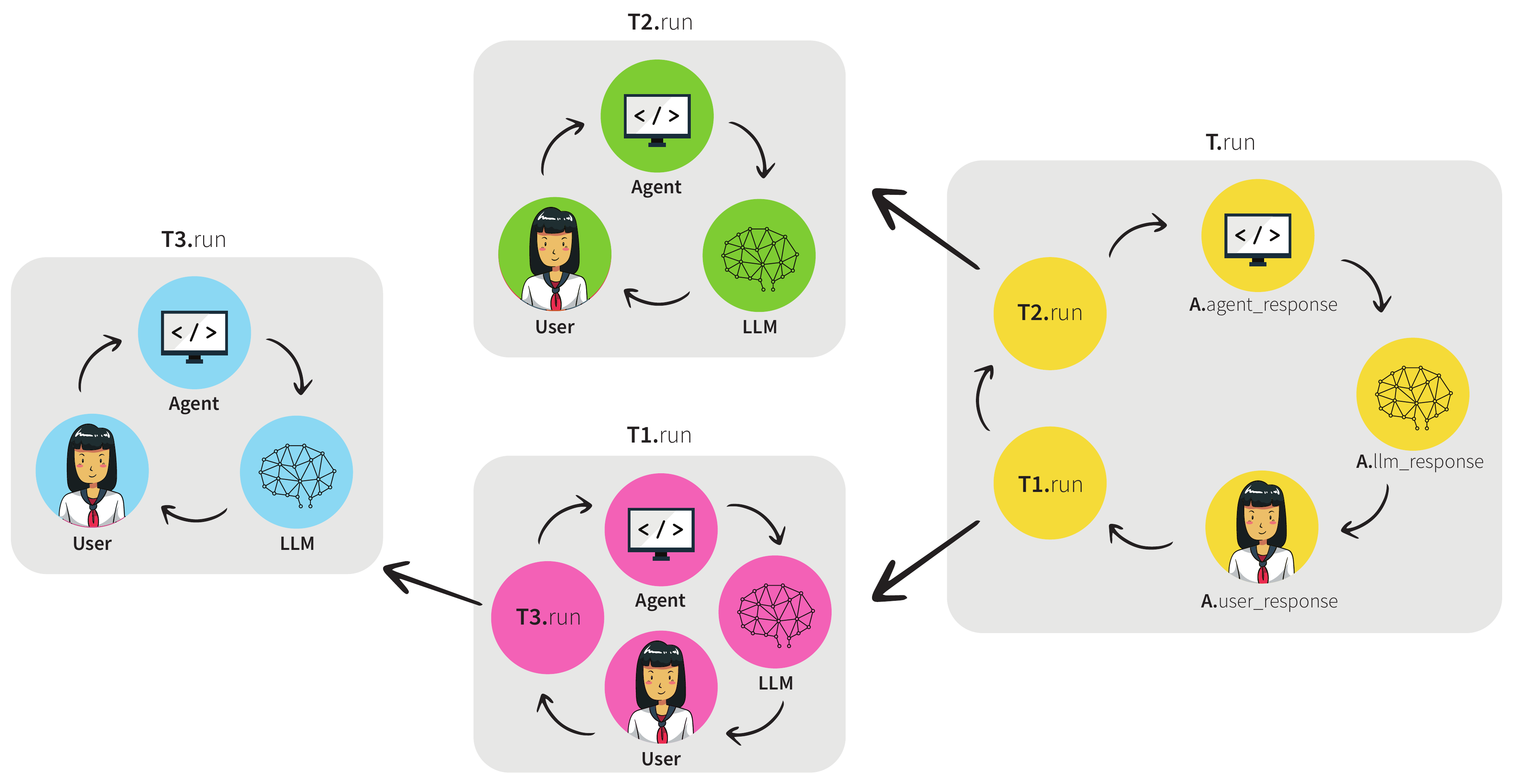}
    \caption{Example of how iteration among responder methods works when a task T has sub-tasks [T1, T2] and T1 has a sub-task T3.}
    \label{fig:app-multiagent}
\end{figure}
As in Figure~\ref{fig:app-multiagent}, the orchestration mechanism is encapsulated in a Task class that wraps an Agent,
and one initiates a task by invoking its \texttt{run} method which has type
signature \stringtostring, identical to the type signature
of an Agent's own native ``response'' methods
(corresponding to the LLM, tool-handler, and human user).
The Task maintains a ``current pending message'' (CPM) to be acted on by one of the ``responders'' of the Task, which includes the agent's own response methods as well as \texttt{run} methods of sub-tasks.
The \texttt{run} method executes a series of ``steps'' until a task termination condition is reached. In each step, a valid response to the CPM is sought by iterating over the responders, and
the CPM is updated with the response.
See Appendix~\ref{app:multi-agent-background} for more details.

\section{\ours: Proposed Multi-Agent System for ADE Extraction}
\label{sec:ours}

In this section, we describe our RAG-based Multi-Agent architecture, \ours, for identifying
associations between drug categories and outcomes.
We first give a high-level outline of the objectives of the key sub-tasks in Section~\ref{sec:ours-outline},
and delve into their implementation details in Section~\ref{sec:ours-agent-critic-pattern} - \ref{sec:ours-step3}.
See Figure~\ref{fig:ours_intro} for an illustrative depiction of the overall pipeline.

{
We emphasize that developing a multi-agent RAG system tailored for ADE extraction
is a highly non-trivial undertaking, requiring careful handling of several issues:
(a) the complex structure of FDA label data, which can be challenging for
naively applied RAG techniques;
(b)
the difficulty of correctly grouping prescribed drugs (\eg assigning the appropriate National Drug Code) based on the varied text descriptions present in medical databases;
(c) LLM brittleness such as deviation from instructions, hallucinations, and inaccurate or incorrect responses;
(d) Infinite loops, fixed points, and deadlocks, which can arise in inter-agent interactions unless carefully managed.
}

\subsection{Objectives of Key Sub-tasks}
\label{sec:ours-outline}

Our ultimate goal is to identify the risk of an adverse event associated with a drug {\em category.}
We developed our system, \ours, to be able to respond to questions of the form:
\begin{quote}
``Does drug category $C$ {\em increases\/} the risk of
a specific (adverse) health outcome $H$, {\em decrease\/} it, or is there {\em no clear effect?}.
And what is the evidence?''
For instance, $C$ could be ``ACE inhibitors'', and $H$ could be ``angioedema''.
\end{quote}
Given a query of this form, the system executes the following steps: given $C$ and $H$,

\begin{adjustwidth}{1cm}{}
\begin{enumerate}[label=STEP {{\arabic*}}:]
    \item Find the extensive list of drug names that belong to $C$ by searching the FDA's National Drug Code (NDC) database.
        Among them, \drugagent identifies drugs $D$ representing each category; top-$k$ distinct drug names that are most commonly prescribed in a clinical database (\eg MIMIC-IV).
    \item For each representative drug $D$ in $C$, \interactionagent generates a free-form (\ie unstructured) text summary about the effect of $D$ on $H$.
    These summaries are generated by referring to up-to-date external pharmaceutical reference sources (\eg FDA drug label database),
    which indicate potential adverse outcomes and evidence for the risks.
    \item \labelagent combines the drug-level information from STEP 2, and generates a structured report; consisting of a label (one of ``increase'', ``decrease'', or ``no-effect'') indicating the potential effect of $C$ on the risk of $H$, a confidence score for this label, structured descriptions of levels of risk, and strength of evidence.
\end{enumerate}
\end{adjustwidth}

Our system extracts the association between $C$ and $H$ by establishing the associations between each drug within $C$ and $H$, rather than directly linking $C$ to $H$.
This construction is motivated by that the reference sources for drug label data, such as the FDA drug label database in our implementation, are typically structured by individual drugs rather than broad drug categories; hence necessitating STEP 1.
It is important to note that applying our system to real patient data requires a complete list of drugs, including both brand and generic names, which can be used to map the actual prescribed drugs recorded in electronic health record (EHR) data to their corresponding categories.

Each of \drugagent, \interactionagent, and \labelagent is coupled with a Critic agent, which provides feedback on the primary agent's output.
The primary agent then regenerates its output based on this feedback.
This Agent-Critic interaction continues until the Critic approves the agent's response.
This design pattern significantly enhances the reliability of our system, as detailed further in Section~\ref{sec:ours-agent-critic-pattern}.

\subsection{Agent-Critic Interaction}
\label{sec:ours-agent-critic-pattern}

This is the core multi-agent interaction pattern that underlies our system,
and is reminiscent of Actor/Critic methods in reinforcement learning~\citep{konda1999actor}.

\paragraph{Agent.}
In an Agent-Critic pair, the Agent is the primary entity that handles external-facing input and output.
It receives a specific goal, instructions on how to accomplish the goal, and access to tools and resources.
In our context, the goal is generally a form of specialized question-answering;
resources can be data sources, or even other agents or multi-agent systems, that the agent
can draw upon when answering the question;
tools are structured responses needed to trigger calls to APIs, database look-ups, or computations.

The primary function of the Agent is to construct a sequence of queries to these resources to fulfill its goal.
The Agent is instructed to compose a semi-structured message consisting of its answer, its reasoning steps and a justification (citing sources where possible) of its answer
in a semi-structured format, and seek feedback on these from the Critic, as below.

\begin{figure}[hb]
    \centering
    \includegraphics[width=\linewidth]{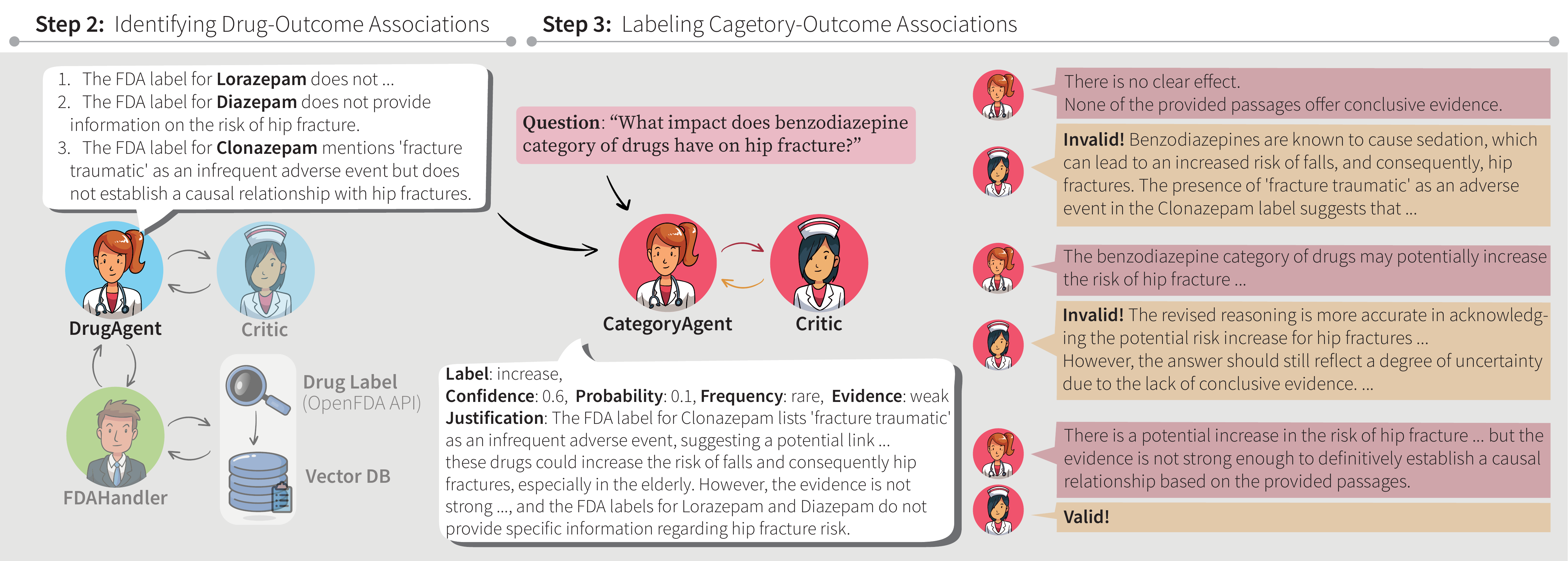}
    \caption{{{\bf Real-world demonstration of Agent-Critic interactions in \ours.}
    Given the question of identifying the association between \textit{Benzodiazepines} and \textit{Hip Fracture}, we illustrate how \labelagent corrects its answers over iterations until the paired Critic is satisfied. See Appendix~\ref{app:agent-critic} for full prompts between the two agents. Agents are instantiated using GPT-4 Turbo.
    }}
    \label{fig:agent-critic}
\end{figure}

\paragraph{Critic.}
This is another agent, paired with the one described above. 
The Critic's role is to validate the Agent's reasoning steps and compliance with instructions, and provide
feedback to the Agent{, which has been shown to improve the quality of LLM-generated outputs~\cite{self-refine}}. The Agent iterates on its response based on this feedback, until the Critic is satisfied,
at which point the Agent signals completion and outputs the results {(see Figure~\ref{fig:agent-critic})}.

While the Agent/Critic pattern may appear simple, this interaction is extremely powerful, and can significantly improve the reliability of the task completion.
This synergistic relationship mirrors a pattern in interactive proof systems used in complexity theory; a prover (\ie Agent) presents a solution, and a verifier (\ie Critic) checks the validity of this solution. The verifier cannot solve the problem on its own but is capable of checking the prover's solution efficiently, which is relatively easier~\citep{babai1985trading}.
This way, even if the Agent's task is complex, the Critic can efficiently verify the correctness of the solution, thereby enhancing reliability.

\subsection{STEP 1: Finding Representative Drugs}
\label{sec:ours-step1}
We first construct a reasonably complete set of all drugs that can possibly belong to the category,
by querying FDA's NDC database, which contains records of specific drugs,
tagged with pharmacological class information of various types (including chemical classes, mechanisms of action, and established pharmacologic classes). Specifically, we extract all drugs
with names or classes matching the relevant search term or terms (\eg ``antibiotic'' or any of the sub-categories
considered by OMOP, for example, erythromycin).
Since this list may contain some drugs that do not actually belong to the class
(\eg a search for ``typical antipsychotics'' returns atypical antipsychotics as well),
we rely on an additional filtering phase to construct the final, reasonably accurate list of
drugs in the category.
For each drug $D$ in this ``complete'' list, we obtain its prescription rate
via a SQL query to the MIMIC-IV prescriptions table.

Note that we chose to implement the above two SQL query steps directly without
using an LLM to generate the queries. This is an instance of an important design principle we
adhere to in our system, which we call the {\em LLM Minimization} principle:
for tasks that can be expressed deterministically and explicitly in a
standard programming paradigm, handle them directly without using LLMs to
enhance reliability and reduce token and latency costs.

\paragraph{\drugagent.}
Now that we have a reasonably complete list of candidate drug names that belong to the category of interest,
along with their prescription rates, we want to identify  three distinct, most commonly prescribed drugs
that belong to the category. This task is complicated by several difficulties:
the same drug may appear in this list with different names;
some pairs of drugs may be essentially the same but only differ in formulation and delivery method, and a judgment
must be made as to whether these are sufficiently different to be considered pharmacologically distinct;
and some of these drugs may not actually belong to the category.
This task thus requires a grouping operation, related to the task of identifying standardized drug codes from text descriptions, well known to be challenging~\citep{10.3389/fbinf.2023.1328613}. Hence, this is very difficult to explicitly
define in a deterministic manner that covers all cases (unlike the above database tasks),
and hence is well-suited to LLMs, particularly those such as GPT-4 Turbo
which are known to have been trained on vast amounts of medical texts in general
(and drug-related ones in particular).
We assign this task to the \drugagent, which is an Agent/Critic system where the Critic agent helps improve the paired agent's output via iterative feedback; in particular, the Critic corrects the Agent when it incorrectly classifies drugs as pharmacologically distinct.

\subsection{STEP 2: Identifying Drug-Outcome Associations}
\label{sec:ours-step2}

\interactionagent is an Agent/Critic system whose task is to identify whether a given drug has an established effect on the risk of a given outcome, based on FDA drug label database, and output a summary of relevant information, including the level of identified risk and the evidence for such an effect. This agent does not have direct access to the FDA Drug Label data, but can receive this information via another agent, \fdahandler.
\fdahandler is equipped with tools to invoke the OpenFDA API for drug label data, and answers questions in the context of information retrieved based on the queries.
Information received from this API is ingested into a vector database, so the agent first uses a tool
to query this vector database, and only resorts to the OpenFDA API tool if the vector database does not contain the relevant information.

\subsection{STEP 3: Labeling Drug Category-Outcome Associations}
\label{sec:ours-step3}

To identify the association between a drug category $C$ and an adverse health outcome $H$,
we concurrently run a batch of queries\footnote{For any OMOP drug categories which contain multiple sub-categories, we execute the full process for each sub-category (identifying a set of representatives for each sub-category), merging the outputs of the classification agent, taking the highest risk indicated for any sub-category as the risk for the full category.}
to copies of \interactionagent, one for each drug $D$ in the category,
of the form: ``Does drug $D$ increase or decrease the risk of condition $H$?''.
The results are sent to \labelagent, described next.

\labelagent is an Agent/Critic system that performs the final classification step;
its goal is to generate a label identifying whether a category of drugs increases or decreases the risk of a condition, or has no effect.
In addition to the label, \labelagent produces a number of additional outputs, all of which are combined into a JSON-structured string{, including}:
(a) a {\em confidence\/} score in [0,1], indicating the confidence in the assigned label,
(c) {\em strength of evidence,} one of ``none'', ``weak'', or ``strong'', and
(d) {\em frequency of the effect,} one of ``none,'' ``rare'', or ``common''.
In this sense, \interactionagent serves as a function of the following type:
\code{[string]\to\{``increase'',``decrease'',``no-effect''\}\cross[0,1]\cross\{``non-\\e'',``weak'',``strong''\}\cross\{``none'',``rare,'',``common''\}}.
The structured output of \labelagent facilitates downstream post-processing to produce a final evaluation,
with no further LLM involvement (Section \ref{sec:omop-setup}).

\section{Experiments}
\label{sec:experimental}
\newcommand{\omoptask}{OMOP ADE task\xspace}

This paper presents \ours, the first LLM-based multi-agent architecture that is capable of producing a structured report with characterizations and scores
related to the risk of an adverse health outcome $H$ from a drug category $C$,
based on FDA drug label data.
We evaluate our method against a widely used benchmark,
the OMOP Evaluation Ground Truth task~\citep{omop}, henceforth referred to as the \omoptask (Section~\ref{sec:omop-setup}), to answer the following three research questions:
\begin{adjustwidth}{1cm}{}
\begin{enumerate}[label=RQ{{\arabic*}}:]
    \item How effectively does \ours identify ADEs? (Section~\ref{sec:rq1})
    \item Does Agent-Critic interaction, the core design pattern underlying \ours, effectively enhance the reliability of the system? (Section~\ref{sec:rq2})
    \item What useful insights do the justifications by \ours provide for further system improvement? (Section~\ref{sec:rq3})
\end{enumerate}
\end{adjustwidth}

\subsection{Evaluation Setup}\label{sec:omop-setup}

The objective of \omoptask is to assign one of three labels (``increase,'' ``decrease,'' and ``no-effect'') to each ($C$, $H$) pair, denoting whether $C$ increases, decreases, or has no effect on the risk of $H$, respectively.
There are 10 drug categories, some of which consist of a single drug, and 10 health outcomes (refer to Table~\ref{tab:drug-outcome-list} for the complete list).
Notably, while only three labels are valid outputs, not all ($C$, $H$) pairs are deemed sufficiently certain to be used in the evaluation.
The authors of \omoptask mark certain pairs as \textit{uncertain}, to which we assign ``no-effect'' labels
with the special restriction that it should not be used in the evaluation. See Appendix~\ref{app:omop-details} for further details.

\begin{figure}[b]
  \centering
  \includegraphics[width=\textwidth]{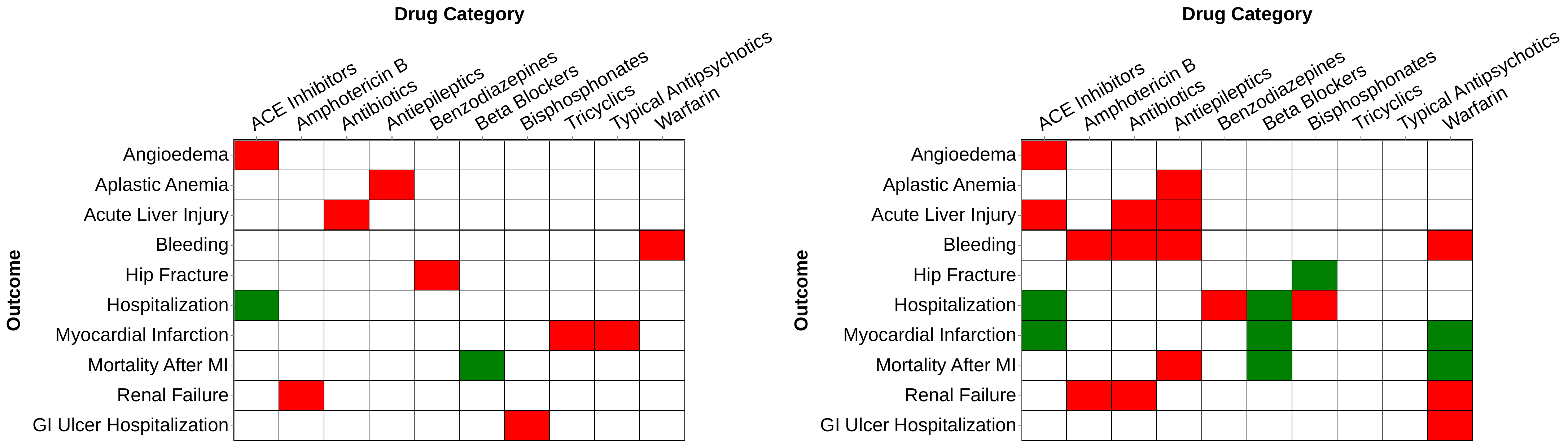}
  \caption{\textbf{Ground truth (left) vs. predictions by \ours (right) for \omoptask.} Red, green, and white cells represent ``increase'', ``decrease'', and ``no-effect'' labels, respectively.}
  \label{fig:omop_results}
\end{figure}

\paragraph{Metrics.}
For quantitative evaluation, we convert the task into binary classification with two different focuses of analysis:
(1) classifying effect vs no-effect, where the labels ``increase'' and ``decrease'' are considered the positive class, and ``no-effect'' is the negative class (namely,  \textbf{effect-based} classification); and
(2) classifying ADE vs. non-ADE, where only ``increase'' is considered the positive class, and the other two labels are the negative class 
(namely, \textbf{ADE-based} classification).
For both choices, we report AUC and F1 scores, which are common evaluation metrics for binary classification~\citep{omop, bayer2021ade}.
Corresponding to the above two binary-classification methods, this results
in 
``effect-based AUC, F1" and  ``ADE-based AUC, F1."
The AUC metric captures how well the scores produced by \ours discriminate  between the classes, while the F1 score assesses the accuracy of the assigned labels in classifying both positive and negative instances.

\subsection{RQ1: \ours effectively identifies ADEs}
\label{sec:rq1}
\begin{figure*}[!t]  
    \centering  
    \label{fig:confusion_matrix_gpt4_turbo}  
    \subfloat[With GPT-4 Turbo~\protect\footnotemark]{  
        \includegraphics[width=0.35\textwidth]{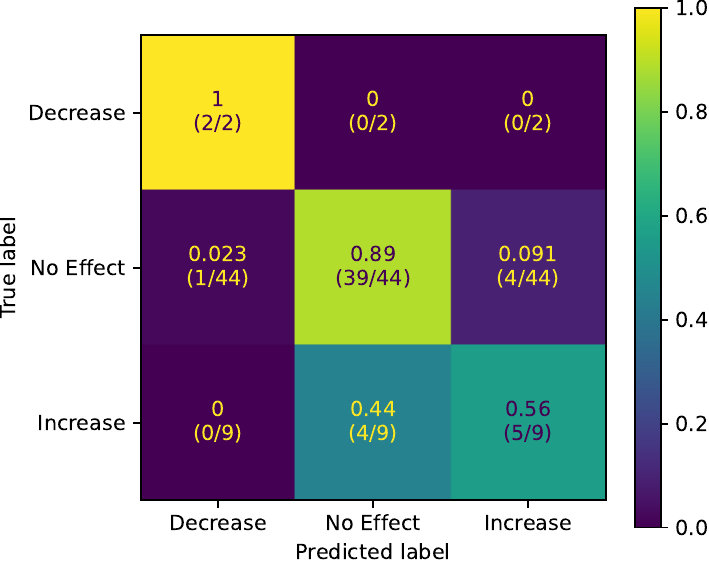}  
    }  
    \label{fig:confusion_matrix_gpt4o}  
    \subfloat[With GPT-4o.]{  
        \includegraphics[width=0.35\textwidth]{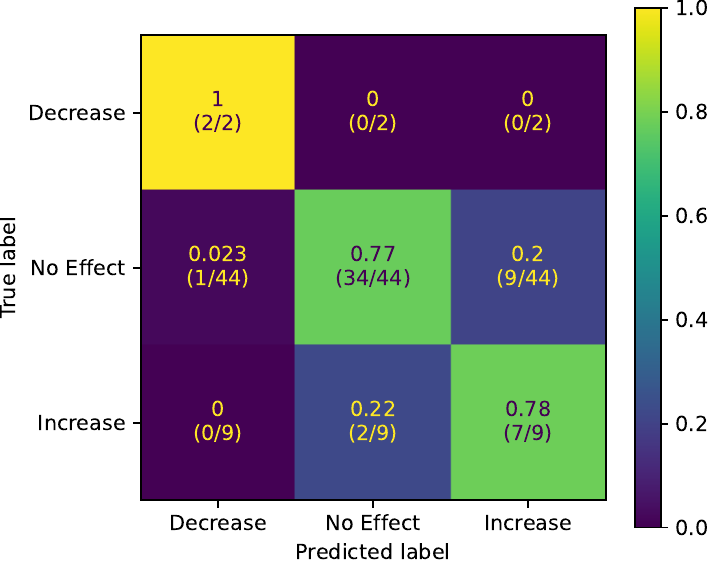}  
    }  
    \caption{\textbf{Confusion matrix for \ours.}}  
    \label{fig:confusion_matrix}  
\end{figure*}

{%
In the evaluations of \ours, we consider two LLMs, GPT-4 Turbo and GPT-4o. For GPT-4o, we limit the number of rounds of feedback from Critics to 5, after which it is required to accept.
Figure~\ref{fig:omop_results} compares the ground truth labels of \omoptask with ADE labels identified by \ours (with GPT-4 Turbo). 
Considering the uncertainty inherent in the label of certain (drug category, outcome) pairs~\citep{omop}, these indicate strong performance on the task.
See Figure~\ref{fig:omop_results_4o} of Appendix~\ref{app:app_experimental} for results on GPT-4o.
We also present the confusion matrix of the \ours labels in Figure~\ref{fig:confusion_matrix}. 

Moreover, we report the performance of \ours in terms of AUC and F1 metrics (see Table~\ref{tab:eval-auc-f1}).
Recall that \labelagent outputs a confidence score ranging from 0 to 1 for its predicted labels, namely "increase," "no-effect," or "decrease.".
This score reflects the agent's certainty regarding the accuracy of the predicted outcome.
For quantitative evaluation as in Table~\ref{tab:eval-auc-f1}, we transform these tripartite label-confidence scores into binary classification probabilities, suitable for effect-based or adverse drug event (ADE)-based analysis. 
Converting the three-class labels to a binary format requires a clear method for correlating each confidence score with a probabilistic value for the respective binary classification task.

The three labels exhibit a natural progression: ``decrease'', ``no-effect'', and ``increase'' imply an \textit{ascending} likelihood that a drug category is associated with the adverse outcome of interest, signifying a rising probability score for the positive class in ADE-based classification.
Furthermore, an \textit{increase} in confidence of "no-effect" or "decrease" corresponds to a \textit{decrease} in the ADE score, while an \textit{increase} in confidence of the "increase" label corresponds to an \textit{increase} in the ADE score.
These observations guide us in formulating an intuitive conversion of the label-confidence scores into ADE probability scores; taking $(1-c_\text{de})/3, (2-c_\text{no})/3,$ and $(2+c_\text{in})/3$, respectively, where $c_\text{de}, c_\text{no},$ and $c_\text{in}$ are the LLM output confidence score when the assigned label is ``decrease'', ``no-effect'', and ``increase'', respectively.
This transformation preserves the semantic ordering of the classes, as well as the valence of confidence in each class. 
To illustrate, increasing confidence in ``decrease'' or ``no-effect'' suggests that the LLM is \textit{less} confident that $C$ causes $H$. 
We derive an effect-score similarly, except that both ``increase'' and ``decrease'' are now positive classes; taking $(1+c_\text{in/de})/2$ and $(1-c_\text{no})/2$, respectively.

The results in Table~\ref{tab:eval-auc-f1} indicate that
  the confidence scores output by the model are well-calibrated.
We observe that \ours performs well both at distinguishing ADEs from non-ADEs and at identifying the presence/absence of an effect in general.
We include ROC curves and sensitivity vs. specificity curves in Figure~\ref{fig:roc} and Figure~\ref{fig:sensitivity_specificity} of Appendix~\ref{app:app_experimental}, respectively. We conduct experiments with additional scoring functions, in particular, the model's estimates of the \textit{probabilities} that $C$ will cause or prevent $H$; see Appendix~\ref{app:probability}.
}

\begin{table}[!t]    
    \centering    
    \begin{adjustbox}{max width=0.7\linewidth}    
        \begin{tabular}{l | l | c | c }      
            \toprule      
            Model & Metric & Effect-based & ADE-based \\ \hline \hline      
            GPT-4o & AUC with confidence & 0.883 & 0.903\\ 
            GPT-4o & F1 score & 0.600 & 0.560\\   
            \midrule      
            GPT-4 Turbo & AUC with confidence & 0.831 & 0.851\\ 
            GPT-4 Turbo & F1 score~\protect\footnotemark[\value{footnote}] & 0.609 & 0.556\\ 
            \bottomrule      
        \end{tabular}    
    \end{adjustbox}    
    \vspace{.2in}
    \caption{{\bf Quantitative evaluation of \ours.} ``Effect-based'' captures the classification between the presence and the absence of any ADE, while ``ADE-based'' represent's the ability of \ours to distinguish drugs with increased risk from those with decreased risk or no effect.}    
    \label{tab:eval-auc-f1}    
\end{table}

\footnotetext{{We observe that GPT-4 Turbo tends to assign ``increase'' rather confidently even when the evidence is weak. To further enhance the reliability of the assigned labels, we take an additional postprocessing step; replacing \textit{unreliable} predictions with ``no-effect''. See Appendix~\ref{app:without-postprocessing} for detailed discussions on label postprocessing.}}

\subsection{RQ2: Agent-Critic interaction enhances reliability}
\label{sec:rq2}

\begin{table}[!htb]  
    \centering %
    \begin{minipage}{.52\linewidth} %
      \centering  
      \begin{adjustbox}{max width=\linewidth}  
      \begin{tabular}{c | c | cc | cc}  
      \toprule  
      \multirow{2}{*}{Critics} & \multirow{2}{*}{RAG} & \multicolumn{2}{c|}{Confidence AUC} & \multicolumn{2}{c}{F1 Score}\\  
      \cline{3-6}  
       & & ADE & Effect & ADE & Effect\\  
      \midrule  
      \checkmark&\checkmark&0.851&0.831&0.556&0.609\\  
      $\times$&\checkmark&0.825&0.819&0.556&0.609\\ \cline{1-6}
      \checkmark&$\times$&0.924&0.929&0.526&0.609\\  
      $\times$&$\times$&0.920&0.926&0.556&0.636\\  
      \bottomrule  
      \end{tabular}  
      \end{adjustbox}
      \vspace{.1in}
      \caption{\bf Ablation results on MALADE.}  
      \label{tab:ablation_main}  
    \end{minipage}%
    \hfill %
    \begin{minipage}{.45\linewidth} %
      \centering  
      \begin{adjustbox}{max width=\linewidth}  
      \begin{tabular}{l | c  }  
      \toprule  
      Agent & Correction\\ \hline \hline  
      \interactionagent & 4.24 \%\\  
      \labelagent & 44.52 \%\\  
      \bottomrule  
      \end{tabular}  
      \end{adjustbox}
      \vspace{.1in}
      \caption{\bf Percentage of agent responses corrected by the Critic.
      }  
      \label{tab:critic1}  
    \end{minipage}  
\end{table}

{%
Our primary tool to analyze the effectiveness of the Agent-Critic pattern in MALADE is by ablation; in particular, we evaluate modified versions of MALADE, with and without feedback from the Critic components of \interactionagent and \labelagent,  with and without RAG for FDAHandler. The results are shown in Table~\ref{tab:ablation_main}.

We observe that, both in the case with and without RAG, Critics improve the quality of the confidence scores, increasing both ADE-based and Effect-based AUCs. We additionally observe strong performance without RAG (in which case Critics slightly improve AUCs but decrease F1 scores), suggesting that GPT-4's internal medical knowledge is frequently sufficient for the \omoptask.
However, to ensure that \ours is a realistic prototype for future pharmacovigilance systems, we consider only instances of MALADE with RAG for our main analysis; LLM-based systems without RAG are prone to hallucinations, and are limited by a static pool of information to draw upon. 
They lack the ability to produce citations, which is vital for trust in these systems, particularly in the medical domain.
Integrating RAG enables the system to access and leverage the most current information from (for example) FDA label data, ensuring the system’s responses are grounded with the up-to-date knowledge available.
Refer to Appendix~\ref{app:ablation} for further details on ablation results and discussions.

We continue our investigation of the effectiveness of Agent-Critic interaction}
by analyzing the frequency of Critic interventions to
rectify errors in Agent responses.
We identify corrections made by the Critic as examples in which the Agent and Critic engaged in more than one round of interaction. Results are shown in Table~\ref{tab:critic1}.

We find that the frequency with which the Critic catches a flaw varies significantly by Agent.
\labelagent in particular incurs errors, necessitating the help of the Critic
and is generally corrected due to flaws in its medical reasoning, hence the Critic can directly prevent an incorrect response.
In the example of an actual run of \ours {(Figure~\ref{fig:agent-critic})}, when asked about the effects of benzodiazepines on hip fracture, \labelagent first answered ``no-effect'',
which was flagged as an error, as the sedative and muscle relaxant properties of benzodiazepines can increase the risk of falls and hence hip fractures, and as \interactionagent had noted that traumatic fractures were listed as an ADE in the drug labels. This feedback was forwarded to \labelagent and used to revise its answer to ``increase''.
We find that \interactionagent generally produces reliable responses; however note that it occasionally makes no calls to the Critic, hence the Agent fails to validate its answer.
We observe that this can occur when the FDA drug label does not contain information related to the condition, and the Agent concludes that no validation is necessary.

\subsection{RQ3: \ours provides justifications that are aligned with human expert reasoning, and help understand its failure modes}
\label{sec:rq3}
We extract the justifications produced by \labelagent from a full run of \ours for \omoptask for review by a clinician.
We observe that the agent exhibits valid medical reasoning in most cases, in particular, $85\%$ of its justifications align with the reasoning of the clinician.

More importantly, examining the provided justifications helps us understand the common patterns of failures and provides guidance on the further improvement of the system.
For instance, \labelagent occasionally assigns ``increase'' to drug categories based on weak evidence, overestimating its strength
It may also overlook risks not explicitly mentioned in the drug label data, particularly when \interactionagent fails to provide sufficient context.
In addition, \labelagent may fail to identify potential therapeutic effects not specified in the drug label data in association with a condition. 
We observe that it does not recognize the antihistamine properties of tricyclic antidepressants. In one case, evidence against gastric and duodenal ulcers caused by alendronate led \labelagent to dismiss results regarding esophageal ulcers.

While \ours exhibits correct medical reasoning in general and hence achieves strong and reliable performance on ADE identification, we highlight that understanding its failures is essential for its further improvements, as discussed in Section~\ref{sec:discussion}. 
Extracts from the logs showing both correct and incorrect behavior by \ours are in Appendix~\ref{app:logs}.
See Appendix~\ref{app:drug_agent_failures} for a discussion of the justifications produced by \drugagent.

\section{Discussion}
\label{sec:discussion}

\paragraph{Generalizable insights about collaborative LLM-powered agents in the context of healthcare.} We have observed the strong performance of \ours for ADE extraction, indicating the potential of multi-agent systems toward broader PhV application. 
Importantly, the principles guiding the design of our system, including 1) the Agent-Critic interaction, 2) the decomposition of a complex task into sub-tasks, and 3) LLM minimization, are quite general.
These principles extend beyond PhV, and can be applied to  many other problems in clinical medicine which require trustworthy, automated responses to challenging questions that must be answered based on multiple competing, and potentially conflicting, sources of knowledge or data. 
Thus, \ours may be viewed not only as a system for ADE extraction, but also as a roadmap for development of other multi-agent systems that generate precise, evidence-based responses to such questions.

\paragraph{General principle 1) Agent-Critic interaction.}
The Agent/Critic pattern, as discussed in Section~\ref{sec:ours}, is essential to the design of our system, and
serves as a powerful tool to enhance accuracy of an LLM-based system. Indeed, we have observed
several instances where the Critic corrected the parent Agent's initial response,
as in the example mentioned in Section \ref{sec:ours-agent-critic-pattern}.
However, we should note that if improperly configured, Critics can be harmful to
the performance of a system, both in terms of efficiency (since the repeated rounds of interaction
between the Agent and Critic can significantly increase token cost and runtime),
and reliability.
Since a Critic strictly enforces the provided guidelines,
incorrect guidelines can significantly harm performance; in some cases, excessively strict requirements can lead to infinite
loops, as the Agent and Critic will deadlock, neither able to satisfy the other's requirements. We observed this
effect in early versions of \ours; resolving the infinite loop issue required specific instructions listing
acceptable behavior. For instance, the Critic for \interactionagent needed to be explicitly told to accept statements
that the effect of a drug was uncertain due to a lack of information from the FDA labels; without this, infinite loops occurred in some drug-outcome combinations.

\paragraph{General principle 2) Decomposition of a complex task.}
The principle of decomposition, mirroring the analogous principle of general software development, is the Unix philosophy as applied to multi-agent systems. Individual agents should be minimal, in that they should
``do one thing and do it well''. This decomposition principle is evident in the hierarchy of specialized agents in the design of \ours (\ie \drugagent, \interactionagent, and \labelagent taking charge of each sub-task in Section~\ref{sec:ours-step1} - \ref{sec:ours-step3}).
In addition to promoting modularity and maintainability, decomposition also promotes reliability,
especially when combined with another key design principle, 
LLM {only when necessary.}

\paragraph{General principle 3) LLM {only when necessary.}}
As LLMs have surprising capabilities, one might be tempted to take an ``LLM-maximalist'' approach, where LLMs are responsible for all aspects of the task.
Unfortunately, this can be both costly and unreliable,
since using proprietary LLMs (\eg GPT-4) behind paid APIs incurs a significant
``token cost'' as well as ``time cost'' (due to the latency of the responses API calls).
Instead, we carefully identified deterministic, well-defined algorithmic parts of the task and
performed these in standard code. 
We relied on LLM-powered agents only for
the specific tasks requiring language understanding, reasoning, and text generation.
This principle guides key choices in \ours: for instance, to retrieve prescription
frequencies of drugs in a category, instead of having an LLM generate the needed SQL queries to the MIMIC-IV database, we observed that these queries are a simple function
of the list of drugs, and directly generated the query in standard code.
A similar choice was made for the FDA API queries to retrieve drug labels.

Such {``LLM only when necessary''} principle also illustrates the key utility of tool-use (also known as function-calling): in addition to providing the LLM the ability to perform external actions and to retrieve external data, it allows offloading execution of complex code from the LLM, hence dramatically reducing cost and increasing reliability. 
A multi-agent orchestration system, in this sense, can be seen as \textit{control flow} for the LLM.

\paragraph{Limitations and Future Work.}
One key limitation of \ours is that we rely entirely on textual FDA label data. In particular, if the information is not specifically included in the label data, \ours cannot reliably identify the strength of any associations raised in the data. This resulted in several flawed predictions, as discussed in Section~\ref{sec:rq3}.
To remedy this, we envision that extracting ADEs from EHR data is a promising direction for future work. As a first step, this would enable estimating the rarity of certain adverse events noted without further detail in the label data; in principle, a multi-agent system with access to EHR data may be able to identify ADEs directly. This would require the LLM to perform causal discovery from historical data (answering, ``Is the drug causing this event?'').

Another interesting avenue for future work is a detailed evaluation with local, open-source LLMs such as LlaMA~\citep{touvron2023llama}, Grok~\citep{grok}, and Mistral~\citep{jiang2023mistral}, which have privacy and cost advantages over the proprietary LLMs (such as GPT-4, Claude, and variants) behind paid APIs.
Unfortunately, our initial experiments with local LLMs exhibited many failure modes due to deviation from instructions and incorrect tool use.
These are in principle possible to remedy by further breaking down tasks into simpler subtasks, and more sophisticated multi-agent validation and correction mechanisms.

Besides these broad limitations and avenues of future work, a few specific improvements are possible. 
Our system requires some minimal human input at the initial step, in particular, the names of the drug categories must be put into the form expected by the FDA's databases; in particular, acronyms are expanded and plurals and punctuation are removed. This task is quite likely amenable to LLMs, which are capable of acronym identification and could attempt additional transformations for more robust output (for example, identifying all synonyms of a pharmacological class; the union of the drugs identified with each search would then be forwarded to \drugagent). In addition, increased usage of structured input and output is a potential enhancement; for example, \interactionagent's reliability might be enhanced by replacing the free-form text output,
using instructions enforcing the presence of certain information, such as the reliability of information and the risk.

\section{Conclusion}
\label{sec:conc}

We consider the problem of ADE Extraction from FDA Drug Labels, a key task in Pharmacovigilance (PhV), and
propose a solution using \ours, based on collaboration among multiple LLM-powered agents equipped
with Retrieval Augmented Generation (RAG).
Our system goes significantly beyond simplistic techniques that only produce a binary label of
presence/absence of association between a drug category and an ADE:
it produces a structured report containing justification for the generated label,
and scores characterizing probability of occurrence, confidence,
strength of evidence, and rarity of the association between a drug category and an ADE.
The scores permit rigorous quantitative evaluation of the system's performance against
the widely-used OMOP Ground Truth table of ADEs, and the results are impressive.
We introduce the agent/critic pattern, a powerful and general design pattern for reliable multi-agent systems.
We hope that our multi-agent architecture and guiding principles will inspire future work on multi-agent approaches to broader PhV and general medical tasks.

\bibliographystyle{plain}
\bibliography{references}

\newpage
\appendix
\section*{\centering Supplementary Material}
{
Section~\ref{app:agent-oriented} includes an in-depth description of the core primitives of our multi-agent framework. 
Section~\ref{app:app_experimental} offers the experimental details, including the system prompts for each agent, the details on our OMOP evaluation, and discussions of the postprocessing of the generated scores and justifications. 
In Section~\ref{app:logs}, we analyze both successful and unsuccessful instances of MALADE, presenting comprehensive logs for selected examples. 
Section~\ref{app:ablation} presents an ablation study that evaluates the individual contributions of key components to the overall system efficacy; namely, the iterative refinement facilitated by Agent-Critic interactions, and the integration of external knowledge through RAG.
Finally, in Section~\ref{app:variance}, we assess how much the variance of numerical outputs by the random sampling of LLMs affects the variance of scores output by the entire MALADE system.
}

\section{Agent-Oriented Programming}
\label{app:agent-oriented}

{This section describes the core abstractions needed to implement a
complex LLM-based application such as \ours. The open-source multi-agent LLM framework
\href{https://github.com/langroid/langroid}{langroid}~\citep{langroid2023}
has an elegant, intuitive and flexible implementation of these abstractions,
and \ours is built on top of this library.}

\subsection{Tool Use: Example}
\label{app:tool-example}

As a simple example, a SQL query tool can be specified as a JSON structure with a \texttt{sql} field (containing the SQL query) and a \texttt{db} field (containing the name of the database).
The LLM may be instructed with a system prompt of the form:
\noindent
\begin{Verbatim}[commandchars=\\\{\}, fontsize=\small]
When the user asks a question about employees, 
use the SQLTool described in the below schema, 
and the results of this tool will be sent back to you, and you can use these 
to respond to the user's question, or correct your SQL query 
if there is a syntax error.
\end{Verbatim}
The tool handler would detect this specific tool in the LLM's response, parse this JSON structure, extract the \texttt{sql} and \texttt{db} fields, run the query on the specified database, and return the result if the query ran successfully, otherwise return an error message.
Depending on how the multi-agent system is organized, the query result or error message may be handled by the same agent (\ie its LLM), which may either summarize the results in narrative form, or revise the query if the error message indicates a syntax error.

\subsection{Retrieval Augmented Generation}
\label{app:rag}

\begin{figure}[t]
   \centering
   \includegraphics[width=\textwidth]{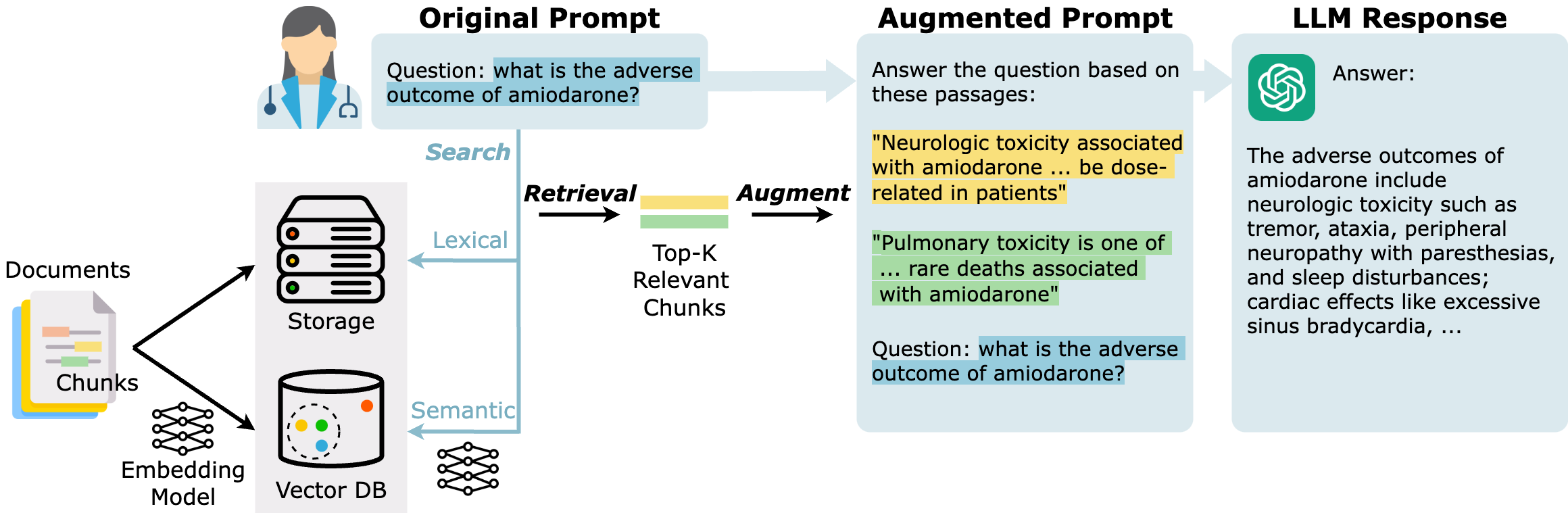}
   \caption{\textbf{A simple agent with RAG}.
   During the ingestion phase, documents are sharded into document chunks.
   At the query phase, top-$k$ chunks most relevant to the original query are retrieved, based on lexical relevance and semantic relevance. Now we prompt the LLM with the augmented query to ground its response to the documents.}
   \label{fig:rag}
 \end{figure}

RAG involves two phases: (a) a \textit{ingestion} phase, where documents are sharded into reasonable-size chunks and ingested into a suitable type of document-store, and (b) a query phase, where top-$k$ document-chunks most \textit{relevant} to the query are retrieved from the document-store, and the LLM is prompted to answer the query given these chunks (see Figure~\ref{fig:rag} for illustrative description).
Not surprisingly, the performance (\ie precision and recall of answers) of a RAG system depends critically on how we define the relevance of document chunks to the query so that they will contain sufficient information for the LLM to compose a reasonable answer.
In this work, we use a combination of two standard notions of relevance: (a) \text{lexical} relevance, which is based on word overlap between the query and the document-chunk (\ie keyword search), while (b) \textit{semantic} relevance focuses on the similarity of ``meaning''. The latter is based on the intuition that specially-trained embedding models
can encode text as fixed-length {\em embedding vectors\/} that roughly capture the ``meaning'' of the text,
and thus two texts are considered semantically similar if their embedding vectors are ``close'' as measured
by a metric such as cosine similarity~\citep{mikolov2013distributed, pennington2014glove, devlin2018bert}.
During the ingestion phase, each document chunk is mapped to an embedding vector using an embedding model and this vector is indexed into a \textit{vector database}, along with a pointer to the chunk contents as metadata.
During the query phase, the same embedding model is used to map the query into a vector, and then the top-$k$ nearest-neighbors of this vector (based on cosine similarity) are found from the vector database, and their corresponding document chunks are retrieved.

\subsection{From LLM to Agent-Oriented Programming}
\label{app:llm-to-agent}
If we view an LLM as a function with signature \stringtostring,
it is possible to express the concept of an agent, tool, and other constructs
in terms of derived function signatures, as shown in
Table~\ref{tab:llm-agent-typesig}.

\newcommand{\inputqry}{\colorbox{cyan!23}{\parbox[c][2.6mm][c]{\widthof{string}}{string}}}
\newcommand{\msghistory}{\colorbox{yellow!55}{\parbox[c][2.6mm][c]{\widthof{[string]}}{[string]}}}
\newcommand{\systemmsg}{\colorbox{pink!55}{\parbox[c][2.6mm][c]{\widthof{string}}{string}}}

\begin{table}[ht]
\resizebox{\linewidth}{!}{
\centering
    \begin{tabular}{l|l}
    \toprule
        Function Description & Function Signature \\ \hline \hline
        LLM & \code{\inputqry\to string} ~~ \code{\inputqry} for the original query.\\ \hline
        Chat interface & \code{\msghistory\cross \inputqry\to string} ~~ \code{\msghistory} is for previous messages\footnote{Note that in reality, separator tokens are added to distinguish messages, and the messages are tagged with metadata indicating the sender, among other things.}. \\ \hline
        Agent & \code{\systemmsg\cross \msghistory\cross \inputqry\to string} ~~ \code{\systemmsg} is for system prompt.  \\ \hline
        Agent with tool & \code{\systemmsg\cross (string\to T)\cross (T\to string)\cross \msghistory\cross \inputqry\to string} \\
        Parser with type \code{T} & \code{string\to T} \\
        Callback with type \code{T} & \code{T\to string} \\ \hline
        General Agent with state type \code{S} & \code{S\cross \systemmsg\cross (string\to T)\cross(S\cross T\to S\cross string)\cross \msghistory\cross \inputqry\to S\cross string} \\
    \bottomrule
    \end{tabular}
}
\caption{\textbf{From LLM to agent-oriented programming.} 
An LLM is essentially a message transformer. Adding ``tool'' (or function calling) capability to LLM requires a parser and a callback that performs arbitrary computation and returns a string.
The serialized instances of \code{T} correspond to a language $L$; as, by assumption, the LLM is capable of producing outputs in $L$, this allows the LLM to express the intention to execute Callback with arbitrary instances of \code{T}. Finally, we incorporate state by making Agent and Callback transducers, and have the general form in the last row.
}
\label{tab:llm-agent-typesig}
\end{table}

\subsection{Detailed Description of Multi-Agent Orchestration}
\label{app:multi-agent-background}

When building an LLM-based multi-agent system, an orchestration mechanism is critical to manage the flow of messages between agents, to ensure task progress, and handle deviations from instructions. In this work, we leverage
{\href{https://github.com/langroid/langroid}{Langroid}}'s
simple yet versatile orchestration mechanism that seamlessly handles:
\begin{itemize}
    \item user interaction
    \item tool handling
    \item sub-task delegation
\end{itemize}

Recall that we view an agent as a message transformer; it may transform an incoming message using one of its three ``native'' responder methods, all of which have the same function signature: \stringtostring:

\begin{itemize}
    \item \llmresponse returns the LLM's response to the input message. 
    Whenever this method is invoked, the agent updates its dialog history (typically consisting of alternating user and LLM messages).
    \item \userresponse prompts the user for input and returns their response.
    \item \agentresponse by default only handles a ``tool message,''  \ie one that contains an llm-generated structured response, performs any requested actions, and returns the result as a string. An \agentresponse method can have other uses besides handling tool messages, such as handling scenarios where an LLM ``forgot'' to use a tool, or used a tool incorrectly, and so on.
\end{itemize}

To see why it is useful to have these responder methods, consider first a simple example of creating a basic chat loop with the user. It is trivial to create such a loop by alternating between \userresponse and \llmresponse. Now suppose we instruct the agent to either directly answer the user's question or perform a web-search. Then it is possible that sometimes the \llmresponse will produce a "tool message", say WebSearchTool, which we would handle with the \agentresponse method. This requires a slightly different, and more involved, way of iterating among the agent's responder methods. From a coding perspective, it is useful to hide the actual iteration logic by wrapping an Agent class in a separate class, which we call a Task, which encapsulates all of the orchestration logic. Users of the Task class can then define the agent, tools, and any sub-tasks, wrap the agent in a task object of class Task, and simply call task.run(), letting the Task class deal with the details of orchestrating the agent's responder methods, determining task completion, and invoking sub-tasks.

The orchestration mechanism of a \Task~ object works as follows. When a task object is created from an agent, a sequence of eligible responders is created, which includes the agent's three ``native'' responder agents in the sequence: \agentresponse, \llmresponse, \userresponse. The type signature of the run is \stringtostring, just like the \Agent's native responder methods, and this is the key to seamless delegation of tasks to sub-tasks. A list of subtasks can be added to a \task~ object via \texttt{task.add\_sub\_tasks([t1, t2, ... ])}, where \texttt{t1, t2, ...} are other \Task~objects. The result of this is that the run method of each sub-task is appended to the sequence of eligible responders in the parent task object.

A task always maintains a current pending message (CPM), which is the latest message "awaiting" a valid response from a responder. At a high level the run method of a task attempts to repeatedly find a valid response to the CPM, until the task is done. This is achieved by repeatedly invoking the step method, which represents a "turn" in the conversation. The step method sequentially tries the eligible responders from the beginning of the eligible-responders list, until it finds a valid response, defined as a non-null or terminating message (i.e. one that signals that the task is done). In particular, this step() algorithm implies that a Task delegates to a sub-task only if the task's native responders have no valid response.

There are a few simple rules that govern how step works: (a) a responder entity (either a native entity or a sub-task) cannot respond if it just responded in the previous step (this prevents a responder from "talking to itself", (b) when a response contains "DONE" the task is ready to exit and return the CPM as the result of the task, (c) when an entity "in charge" of the task has a null response, the task is considered finished and ready to exit, (d) if the response of an entity or subtask is a structured message containing a recipient field, then the specified recipient task or entity will be the only one eligible to respond at the next step.

Once a valid response is found in a step, the CPM is updated to this response, and the next step starts the search for a valid response from the beginning of the eligible responders list. When a response signals that the task is done (e.g. contains the special string "DONE"), the run method returns the CPM as the result of the task. This is a highly simplified account of the orchestration mechanism, and the actual implementation is more involved.

The above simple design is surprising powerful and can support a wide variety of task structures, including trees and DAGs. As a simple illustrative example, tool-handling has a natural implementation. The LLM is instructed to use a certain JSON-structured message as a tool, and thus the \llmresponse method can produce a structured message. This structured message is then handled by the \agentresponse method, and the resulting message updates the CPM. The \llmresponse method then becomes eligible to respond again, and the process continues.

Figure~\ref{fig:app-multiagent} shows a schematic of the task orchestration and delegation mechanism.

\section{Detailed Descriptions on \ours Implementation}
\label{app:app_experimental}

\subsection{Prompts to Each Agent}
\label{app:prompt}

\paragraph{STEP1: finding representative drugs under each drug category.}
This is the full prompt to \drugagent:
\begin{lstlisting}
    You are a helpful assistant with general medical and pharmacological knowledge.  I will provide you with a list of drugs, and the result of a query on a medical database with their usage rates; your goal is to find N representative drugs in category \{cat\} out of the provided drugs.

    Prefer generic names if possible, and do not include both a brand and generic name for the same drug in your list.

    If possible, prefer drugs with different active ingredients 
    (i.e. avoid derivatives of a drug already in the list), 
    keeping your choices to the most basic variant of a given drug 
    from the list (use the total prescription rate of variants of the same base drug to select the top drugs); disregard this if you cannot find N with this restriction. If fewer than N meet the conditions, you may include fewer than N (but never more).

    The names of the selected representatives must EXACTLY match one of the provided drugs; choose the names from the original list, not the database query.

    You must provide your final answer with the `final_answer` tool/function; make sure to clearly state my question, as well as the reasoning used to derive the answer. Include the requirements on your answer in the `question` field.

    Once the critic is satisfied with your answer, send me the answer with the `submit_answer` tool/function.
\end{lstlisting}

This is the full prompt to the Critic agent:
\begin{lstlisting}
    You are also an expert in medical and pharmacological reasoning.

    Your goal is to ensure that the selected drugs are distinct members of the category \{cat\} of drugs. You will consider information provided directly to the user to be reliable (for example, this might include prescription rates and a complete list of drugs in category \{cat\}). Unless this contradicts your pharmacological knowledge, the user's choices of representatives for a category are acceptable unless they do not represent the basic form of a given drug.
\end{lstlisting}

\paragraph{STEP2: identifying the interaction between each drug and each outcome.}

Below is the full prompt to \interactionagent.
\begin{lstlisting}
    You will receive questions involving medical data.  
    You are experienced in general medical reasoning, but must consult references for any specific medical knowledge required to answer my questions.

    You have access to `FDAHandler`, who will answer questions you ask about specific drugs using FDA data. You must use the `recipient_message` tool/function to ask these questions, and the `intended_recipient` MUST be `FDAHandler` anytime you use this tool. 
    Ensure that you ask FDAHandler for the specific information you need.

    As some potential complications are listed in FDA labels as lacking a verified causal relationship, make certain that your final answer expresses the degree of reliability of your answer. Similarly, make sure to clearly express the degree of risk associated  (i.e. is the condition a rare or a common side effect, or does a drug rarely or frequently result in reduced risk of a condition).

    If FDAHandler cannot answer your question then your answer
    should be {NO_ANSWER}, because the FDA label data does not
    specify the answer. If FDAHandler answers with {NO_ANSWER}
    that means that the FDA label for the drug does not
    contain the information requested (and, in particular, it
    means that it does not mention the condition); hence, your
    answer should be {NO_ANSWER}. This indicates that there
    may not be any effect on the risk of the condition, make sure to  explain this in your justification.

    IMPORTANT: if multiple attempts fail to retrieve any relevant information, there is no need to continue asking questions to FDAHandler; assume that the information is not in the FDA labels and so FDAHandler cannot answer.

    You MUST specifically tell the critic why you could not
    find an answer to the question; be sure to specify that
    the FDAHandler answered with {NO_ANSWER} if that is the reason.

    You must provide your final answer with the `final_answer`
    tool/function; make sure to clearly state my question, the
    reasoning used to derive the answer, including the questions asked to FDAHandler and a summary of the results, as well as your final answer in the `answer` field.

    Once the critic is satisfied with your answer, say {DONE},
    and give me the answer and justification for it. Make sure
    to provide your answer again, do not just use the answer
    sent to the critic. Include any relevant details provided by FDAAgent.

    If the critic is satisfied and your answer is {NO_ANSWER},
    say {DONE} {NO_ANSWER} and provide a justification.
    IMPORTANT: say {DONE} specifically, not DONE.
\end{lstlisting}

This is the full prompt to the Critic agent:
\begin{lstlisting}
    You are also experienced in medical reasoning, and have general medical knowledge.  Unless the responses are inconsistent with your medical (or common-sense) knowledge,  you generally trust responses from FDAHandler.

    The answer should express the strength of evidence for the answer and the magnitude of the effect.  If the user states that FDAAgent does not have this information, you should accept it.

    If the answer given contains {NO_ANSWER}, accept it as long as the answer clearly expresses why it was not possible to answer the question. If it states that this is because FDAHandler responded with {NO_ANSWER}, you should accept it as sufficient justification.
    Otherwise, ask the user to express whether FDAHandler responded with {NO_ANSWER}, and, if not, to state why it was not possible to answer the question. If it does so, the answer is acceptable and the other requirements need not be enforced.
\end{lstlisting}

In this case, the Critic agent similarly behaves as a medical expert; in general, the Critic must always behave as if proficient with any task that the orchestrator agent will do; this is specified as: ``You are also experienced in medical reasoning, and
            have general medical knowledge. Unless the responses
            are inconsistent with your medical (or common-sense)
            knowledge, you generally trust responses from
            FDAHandler.''

It is told to trust the agents' responses as any necessary validation of the responses from the two agents should happen on their side; the criticism should focus on the orchestrator itself.

Below is the full prompt to \fdahandler:
\begin{lstlisting}
    You will try your best to answer my questions, in this order of preference:

    1. Ask me for some relevant text, and I will send you. 
        Use the `relevant_extracts` tool/function-call for this purpose. 
        Once you receive the text, you can use it to answer my question. 
        If the question asks for information about a specific drug, make sure to begin by including that  drug in the `filter_drugs` field.  If I say {NO_ANSWER}, it means I found no relevant docs, and you can try the next step, using a web search.
    2. If you are still unable to answer, you can use the `relevant_search_extracts` tool/function-call to get some text from a web search. Once you receive the text, you can use it to answer my question. If you need to identify the drugs in a category, use the `drug_category_search` tool/function-call instead.
    3. If you are still unable to answer, and used `filter_drugs` in your initial attempt with `relevant_extracts`, try again without a filter.
    4. If you still can't answer, simply say {DONE} {NO_ANSWER} 

    If given a question asking about a drug "X and Y", this is a
    combination drug, so your initial searches should be for "X and Y" not "X" or "Y".

    If asked a question about drugs in broad category, make to consider EVERY drug in the category, and in particular, if the question asks for which drugs in the category something is true, make CERTAIN that your answer correctly lists ALL drugs in the category where the condition holds.

    IMPORTANT: some fields in the FDA label data retrieved 
    by `relevant_search_extracts' and `relevant_extracts` have the level of reliability  of information specified prior to it (for example, statements of the level of reliability may precede each section of adverse reactions, the immediately preceding such statement is the one that corresponds to any given reported interaction). Make certain that your answer reflects the specified level of reliability. 
    Similarly, when asked about the effect of a drug on a condition,  ALWAYS express the magitude of the effect (i.e. how frequently the drug results in the condition or how frequently the drug improves the condition); whenever possible, make sure to explicitly state whether a condition is rarely or commonly reported.

    ANSWER FORMAT:

    ALWAYS present your answer in one of the below 2 formats:

    1. In case you COULD NOT find an answer:

    {DONE} {NO_ANSWER}  

    2. In case you ARE able to find an answer: 

    {DONE}
    ANSWER: [Your concise answer, with a brief summary of necessary context. ALWAYS clarify the level of reliability of the information, if specified in the extracts. If applicable, ALWAYS express the magnitude of any increase or decrease in risk and any associated information.]
    SOURCE: aspirin label
    EXTRACT_START_END: Aspirin can cause ... with any medicine.

    For the EXTRACT_START_END, ONLY show up to the first 3 words and last 3 words.
\end{lstlisting}

\paragraph{STEP3: labeling the association between each drug category and each outcome.}

This is the full prompt to \labelagent:
\begin{lstlisting}
    You are experienced in general medical reasoning and have general medical knowledge.

    You will be provided a list of passages answering, for each of a set of drugs X, whether drug X increases or decreases the risk of {condition}. They all belong to category {cat_name}.

    You must provide your final answer with the `final_answer` tool/function;  make sure to clearly state my question, the reasoning used to derive the answer, 
    including the evidence from the passages,  as well as your final answer in the `answer` field.

    Once the critic is satisfied, submit your answer with the `category_effect` tool,  making sure that the answer, `label`, is one of the following: "increase," "decrease," or "no-effect," and make sure to include your justification. DO NOT use this tool before you have used the `final_answer` tool and have had your answer accepted by the critic.

    Your `justification` must clearly express the magnitude of risk indicated and the strength of evidence.  Provide a `confidence` value between 0 and 1 indicating the confidence in your assigned `label` and a `probability` value indicating the probability that the drug will cause the condition (or prevent the condition) in a given patient.

    Express the frequency that the drug has an effect as either "none," "rare," or "common" with the `frequency` field and express the strength of `evidence` as either "strong"  (for example, evidence is strong  when shown in a cal trial) or "weak" (for example, this applies to  purely correlational evidence) or  "none" if no evidence exists.
\end{lstlisting}

This is the full prompt to the Critic agent:
\begin{lstlisting}
    You are also experienced in medical reasoning, and have general medical knowledge.  Unless the responses are inconsistent with your medical (or common-sense) knowledge, you generally trust responses from FDAHandler.  Similarly, you trust that the user's statements about passages are correct without the need to review them directly.

    The answer provided should indicate an increase, decrease, or no effect on the risk, and must be no effect if no evidence linking the drug category to the risk of the condition exists.

    The answer should be drawn from the specified passages, hence, the absence of information related to a condition in the FDA data for all drugs in a category should be enough to conclude that there is no effect for that drug.

    The answer should express the degree of certainty and the magnitude of change in risk, ensure that the provided answer is consistent with the evidence.
\end{lstlisting}

\label{app:probability}
\subsection{ Probability-based scoring}

In addition to the confidence-based scoring discussed in Section~\ref{sec:experimental}, we consider probability-based scoring. In particular, we ask the model to specify the probability of an evant, specifically, the event that a drug in category $C$ causes or prevents $H$.

\begin{figure}[h]
  \centering
  \includegraphics[width=0.45\textwidth]{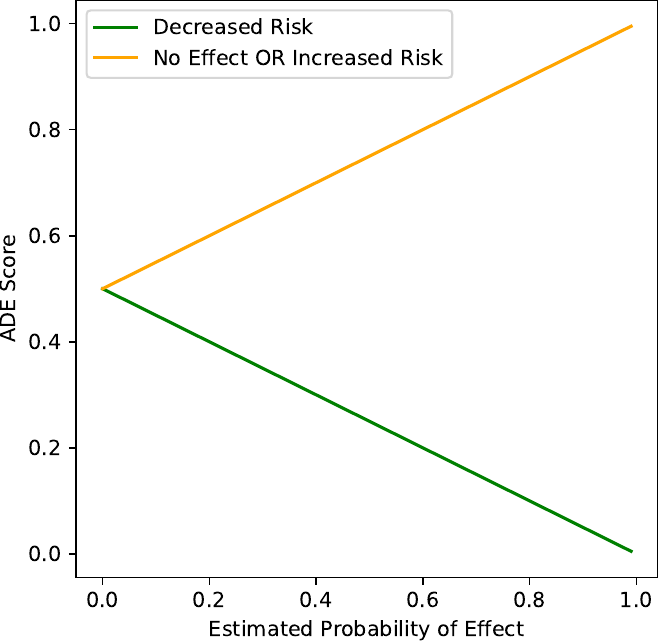}
  \caption{\textbf{Derivation of ADE Scores from Event Probability.} The x-axis represents the LLM's output probability estimate, and the color indicates the mapping for the corresponding set of labels.}
  \label{fig:probability_scores}
\end{figure}

As in Section~\ref{sec:experimental}, we must derive confidence scores in ADE and effects from the output probability estimate; as the probability is already in terms of the probability of any effect, either harmful or beneficial, we use the probability directly for Effect AUC. For ADE AUC, we use the tranformation shown in Figure~\ref{fig:probability_scores}.

We make the assumption that, if the LLM specifies a probability $p$ with a label other than ``decrease,'' that probability expresses the probability of a harmful effect. Hence, the derived score decreases linearly with increasing probability when the label is ``decrease,'' and increases linearly when the label is anything else. With this assumption, we additionally maintain the semantic ordering of the LLM's implied confidence in ADE, and hence this is a well-defined confidence score.

\begin{table}[!t]    
    \centering    
    \begin{adjustbox}{max width=0.6\linewidth}    
        \begin{tabular}{l | l | c | c }      
            \toprule      
            Model & Metric & Effect-based & ADE-based \\ \hline \hline      
            GPT-4o & AUC with confidence & 0.8833 & 0.9034\\      
            GPT-4o & AUC with probability & 0.6715 & 0.6534\\      
            \midrule      
            GPT-4 Turbo & AUC with confidence & 0.8306 & 0.8514\\      
            GPT-4 Turbo & AUC with probability & 0.8058 & 0.7935\\      
            \bottomrule      
        \end{tabular}    
    \end{adjustbox}    
    \vspace{.2in}
    \caption{{\bf Comparison of confidence and probability based scoring for \ours.} ``Effect-based'' captures the classification between the presence and the absence of any ADE, while ``ADE-based'' represent's the ability of \ours to distinguish drugs with increased risk from those with decreased risk or no effect.}    
    \label{tab:eval-auc-probs}    
\end{table}  

The results with probability-based scoring are shown in Table~\ref{tab:eval-auc-probs}, we observe that the probabilities are less reliable (unsurprising as the FDA label data does not always contain the information necessary for a reliable estimate).
  See Appendix~\ref{app:ablation} for further discussion on the potential unreliability of the probability estimates.

\begin{table}[b]
\resizebox{\textwidth}{!}{
\centering
    \centering
    \begin{tabular}{c|c}
    \toprule
        \multirow{3}{*}{Drug Categories} & ACE Inhibitors, Amphotericin B, 
        Antibiotics (Erythromycin, Sulfonamide, Tetracycline), \\
        & Antiepileptics (Carbamazepine, Phenytoin),  Benzodiazepines, Beta blockers, \\
        & Bisphosphonates (Alendronate), Tricyclic antidepressants, Typical antipsychotics, Warfarin
        \\ \cline{1-2}
        \multirow{3}{*}{Outcome} & Angioedema, Aplastic anemia, Acute liver injury, Bleeding, Hip fracture, \\ 
        & Hospitalization, Myocardial infarction, Mortality after myocardial infarction, \\
        & Renal failure, Gastrointestinal ulcer hospitalization\\
    \bottomrule
    \end{tabular}
}
\caption{OMOP drug categories and conditions. Parenthesized lists contain the subcategories of the broad drug category considered.}
\label{tab:drug-outcome-list}
\end{table}

\subsection{\omoptask Details}
\label{app:omop-details}

\begin{figure}[h]
  \centering
  \includegraphics[width=\textwidth]{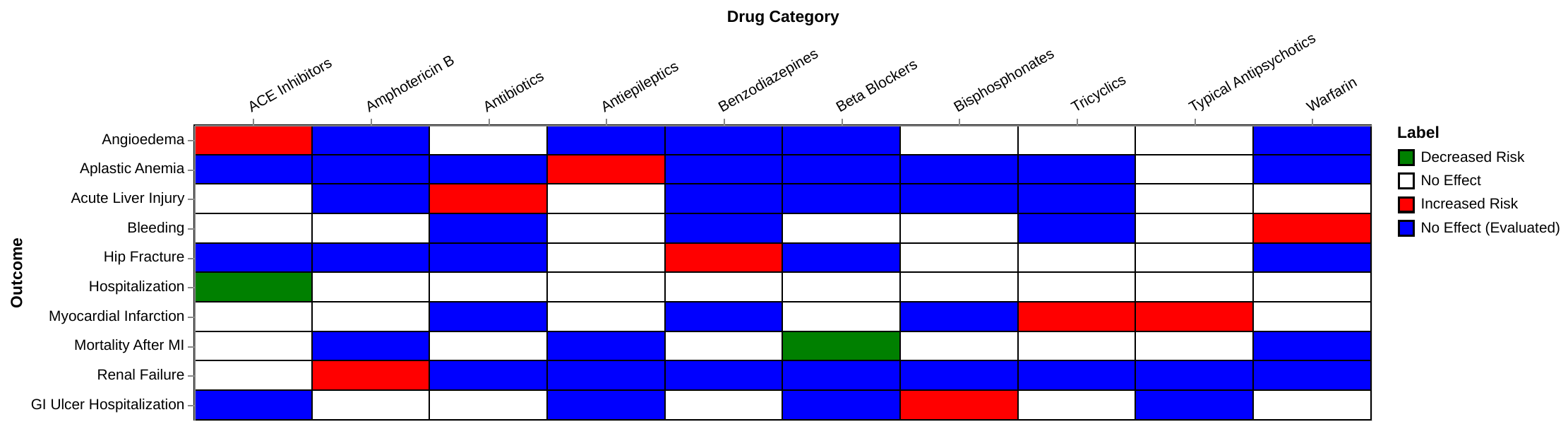}
  \caption{OMOP ground truth.}
  \label{fig:omop_ground_truth}
\end{figure}

The ground truth for the \omoptask is shown in Figure~\ref{fig:omop_ground_truth}. As noted in Section~\ref{sec:omop-setup}, while the \omoptask permits only three output labels for the effect of a drug category on an outcome, some drug category, outcome pairs are considered uncertain (which we treat as a ``no-effect'' label which is not used in evaluation). In Figure~\ref{fig:omop_ground_truth} the ``No Effect'' cells considered reliable are shown in blue, while the uncertain ``no-effect'' cells are those in white. In particular, the cells in white are \textit{not} used in evaluation (\ie for AUC computation, confusion matrices, and F1 scores). The cells used for evaluation are shown in red, blue, and green.

Table~\ref{tab:drug_representatives} present the representatives selected for each category of drugs, respectively, produced by GPT-4 Turbo.
\begin{table}[h]
    \centering
    \begin{tabular}{l | l}
    \toprule
        Drug or Drug Category & Representative Drug(s)\\
        \midrule
        ACE Inhibitors & Lisinopril, Captopril, and Enalapril Maleate\\
        Amphotericin B & Ambisome, Amphotericin B, and Abelcet\\
        Erythromycin & Erythromycin, Erythromycin Ethylsuccinate, and Erythromycin\\
        Sulfonamides & Silver Sulfadiazine, Bactrim, and Sulfadiazine\\
        Tetracyclines & Doxycycline Hyclate, Tigecycline, and Minocycline\\
        Carbamazepine & Carbamazepine\\
        Phenytoin & Phenytoin Sodium, Phenytoin, and Extended Phenytoin Sodium\\
        Benzodiazepines & Lorazepam, Diazepam, and Clonazepam\\
        Beta Blockers & Metoprolol Tartrate, Labetalol, and Atenolol\\
        Alendronate & Alendronate Sodium and Alendronate\\
        Tricyclics & Doxepin HCL, Desipramine, and Amitriptyline HCL\\
        Typical Antipsychotics & Haloperidol, Thiothixene, and Pimozide\\
        Warfarin & Warfarin\\
        \bottomrule
    \end{tabular}
    \caption{Drug Representatives selected for each OMOP category (or subcategory).}
    \label{tab:drug_representatives}
\end{table}

\subsection{Effectiveness of Label Postprocessing {with GPT-4 Turbo}}
\label{app:without-postprocessing}
In this subsection, we illustrate that the postprocessing of labels in Section~\ref{sec:omop-setup} significantly improves the accuracy of ADE identification by \ours instantiated with GPT-4 Turbo.

We take an additional postprocessing step to further enhance the quality of the assigned labels, replacing \textit{unreliable} predictions with ``no-effect,'' unless stated otherwise.
Specifically, we consider outputs for which the LLM reported weak evidence and rare incidences of effects as unreliable. Additionally, we deem outputs for which the LLM selected small round numbers for the probability (\ie 0.1 and 0.01) as unreliable, as such values are often chosen in the absence of strong evidence, resembling typical human preferences for round numbers. We apply this postprocessing except in the case of AUC, as uncertainty should be reflected directly in the confidence scores.

\begin{figure*}[!t]  
    \centering
    \label{fig:confusion_matrix_gpt4_turbo_no_postprocessing}
    \subfloat[Confusion matrix for \ours]{  
        \includegraphics[width=0.33\textwidth]{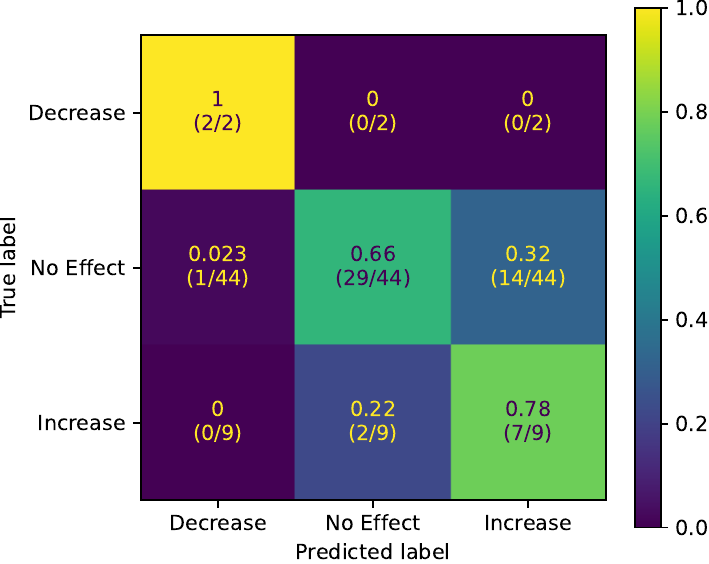}  
    } \\
    \label{fig:omop_results_turbo_no_postprocessing}
    \subfloat[Predictions of \ours]{  
        \includegraphics[width=\textwidth]{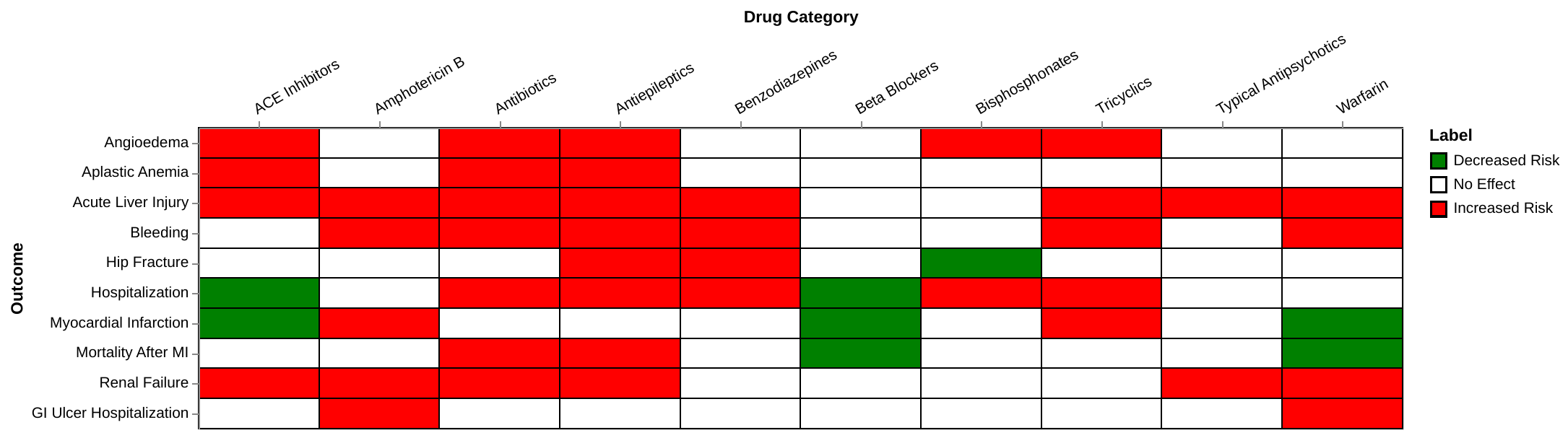}  
    }  
    \caption{Results with GPT-4 Turbo, without postprocessing.}  
    \label{fig:results_turbo_no_postprocessing}  
\end{figure*}

We obtain an effect-based F1 score of 0.5294 without postprocessing, and 0.6087 with postprocessing.
We obtain an ADE-based F1 score of 0.4828 without postprocessing, and 0.5556 with postprocessing.
Figures~\ref{fig:confusion_matrix_gpt4_turbo_no_postprocessing} and~\ref{fig:omop_results_turbo_no_postprocessing} show the confusion matrices and predictions, respectively, of \ours without postprocessing on GPT-4 Turbo (compared to the results in Section~\ref{sec:rq1}).

\begin{figure}[t]
  \centering
  \includegraphics[width=\textwidth]{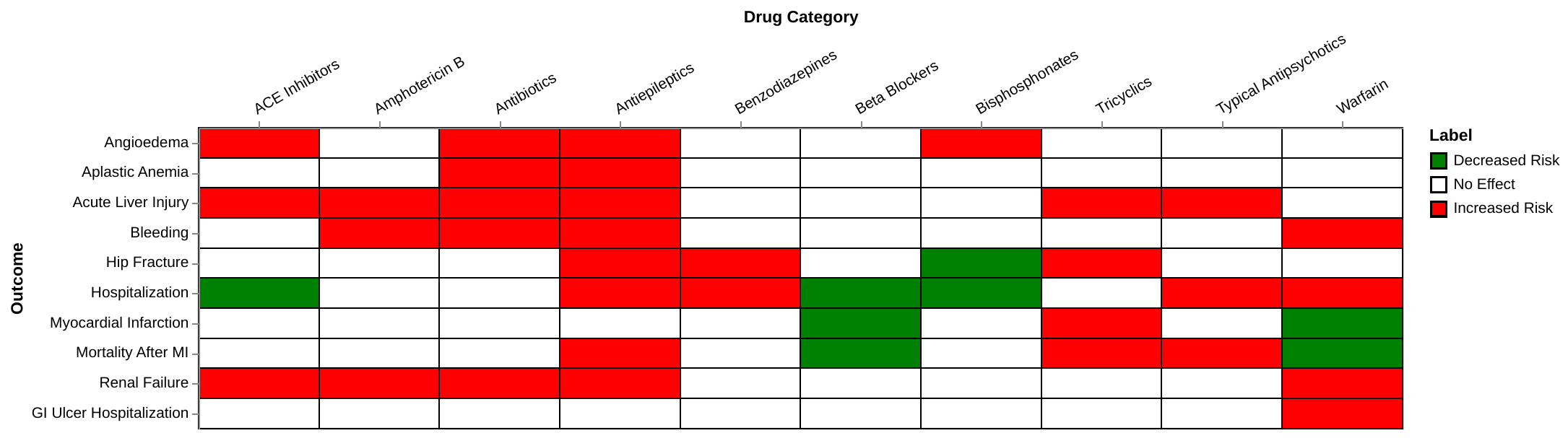}
  \caption{Predictions of \ours run on GPT-4o.}
  \label{fig:omop_results_4o}
\end{figure}

\begin{figure*}[!t]
    \centering
    \subfloat[Using the model's generated confidence values.]{
        \includegraphics[width=.7\textwidth]{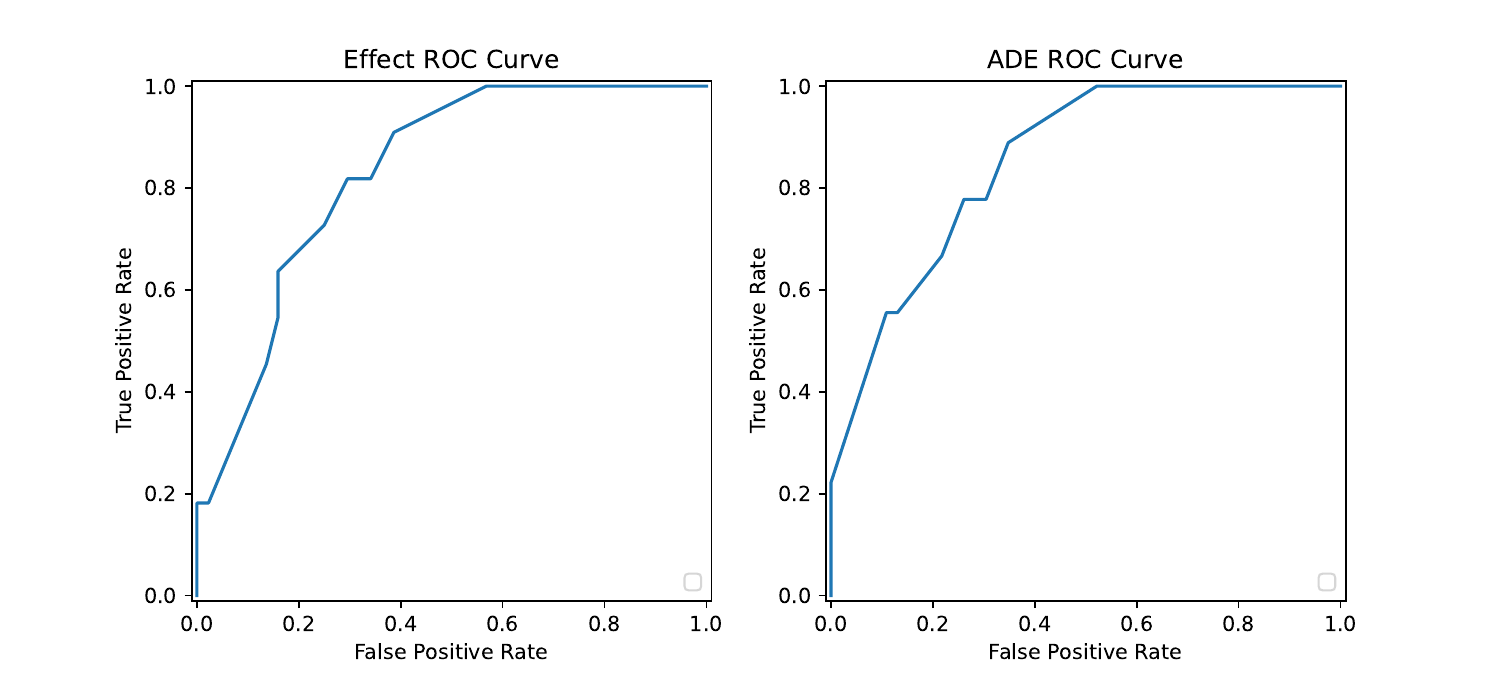}
    } \\
    \subfloat[Deriving confidence from the model's generated probability values.]{
        \includegraphics[width=.7\textwidth]{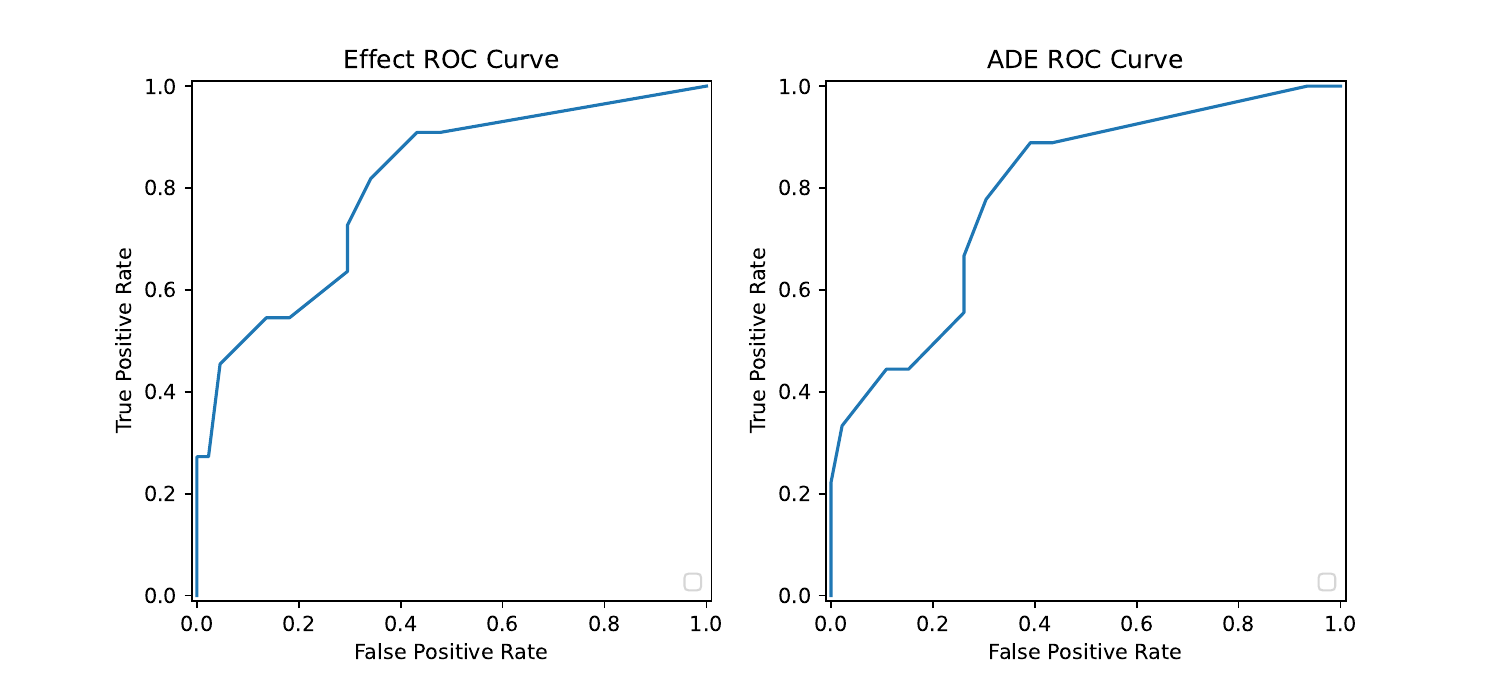}
    }

    \caption{\textbf{ROC curves for \ours on OMOP}}
    \label{fig:roc}
\end{figure*}

\begin{figure*}[!t]
    \centering
    \subfloat[Using the model's generated confidence values.]{
        \includegraphics[width=.7\textwidth]{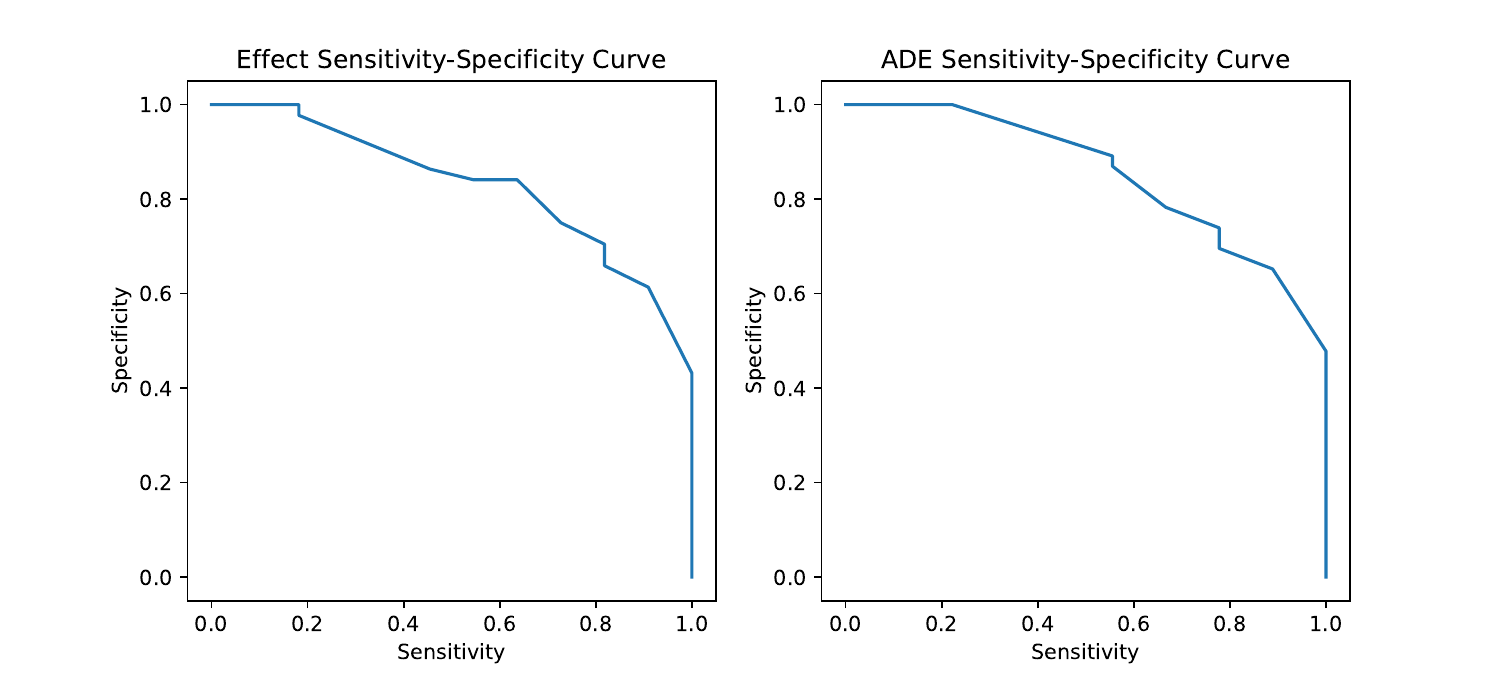}
    } \\
    \subfloat[Deriving confidence from the model's generated probability values.]{
        \includegraphics[width=.7\textwidth]{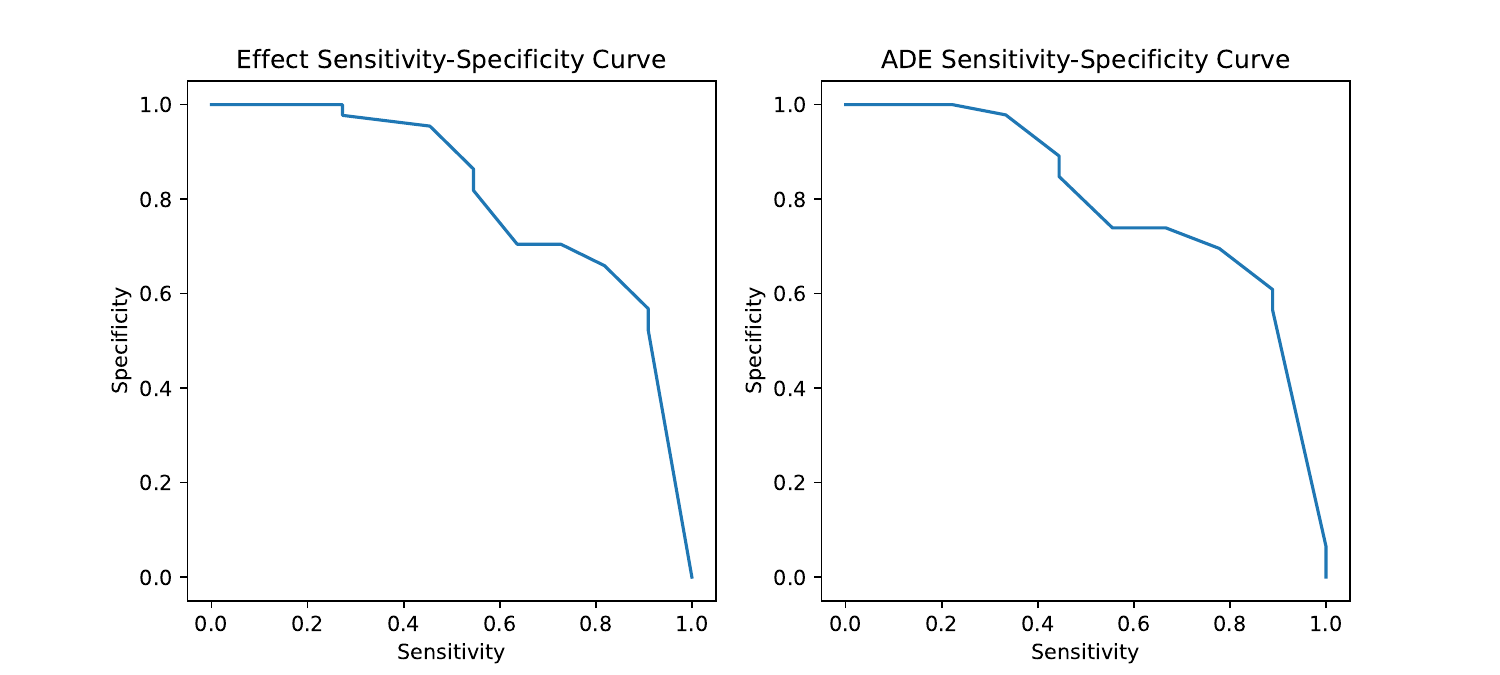}
    }

    \caption{\textbf{Sensitivity vs. specificity curves for \ours on OMOP}}
    \label{fig:sensitivity_specificity}
\end{figure*}

\subsection{Discussion of Justifications Produced by \drugagent}\label{app:drug_agent_failures}

As in Section~\ref{sec:rq3}, we extract the justifications produced by \drugagent in a run of \ours for review by a clinician. The most common flaw in its reasoning is that \drugagent frequently fails to recognize OMOP categories which consist of a single drug; beyond that, the agent's justifications are generally correct, with only one significant other error occurring.

This error isn't directly caused by \drugagent, but, rather, was a result of our OpenFDA retrieval process failing to identify certain common typical antipsychotics, for example, fluphenazine is labeled as a phenothiazine in the National Drug Code DB but not as a typical antipsychotic; hence a more robust search, taking subcategories into account, for drugs in a category is needed.

\section{Detailed behavior of \ours: Successful and Failed Behavior}
\label{app:logs}

\subsection{Successful Behavior}

For an example of a successful run of \ours, consider the effect of ACE inhibitors on agioedema. As we will see, \interactionagent will retrieve the reports of ADEs from the FDA labels, and \labelagent will conclude that they are strong evidence for increased risk.

\subsubsection{\interactionagent Step}

This step will be run three times, for each of the representative drugs (Lisinopril, Captopril, and Enalapril Maleate). We'll show the results for Lisinopril.

The original question to \interactionagent is ``Does LISINOPRIL increase or decrease the risk of angioedema?''

As \interactionagent requires evidence from the drug labels to answer the question, the agent immediately asks \fdahandler, specifying it as the \textit{recipient} of the message using a tool call, as follows:
\begin{Verbatim}[commandchars=\\\{\}, fontsize=\small]
FUNC: \{
  "name": "relevant_extracts",
  "to": "",
  "arguments": \{
    "query": "Does LISINOPRIL increase the risk of angioedema?",
    "filter_drugs": [
      "Lisinopril"
    ]
  \}
\}
\end{Verbatim}

The agent searches its vector DB for relevant passages from the drug's label, using another tool call:
\begin{Verbatim}[commandchars=\\\{\}, fontsize=\small]
FUNC: \{
  "name": "relevant_extracts",
  "to": "",
  "arguments": \{
    "query": "Does LISINOPRIL increase the risk of angioedema?",
    "filter_drugs": [
      "Lisinopril"
    ]
  \}
\}
\end{Verbatim}

A large number of passages are retrieved, including:
\begin{Verbatim}[commandchars=\\\{\}, fontsize=\small]
LISINOPRIL: warnings: WARNINGS General Lisinopril Anaphylactoid and
 Possibly Related Reactions: Presumably because angiotensin-converting
 enzyme inhibitors affect the metabolism of eicosanoids and
 polypeptides, including endogenous bradykinin, patients receiving ACE
 inhibitors (including lisinopril and hydrochlorothiazide tablets) may
 be subject to a variety of adverse reactions, some of them
 serious. Head and Neck Angioedema: Angioedema of the face,
 extremities, lips, tongue, glottis and/or larynx has been reported
 rarely in patients treated with angiotensin converting enzyme
 inhibitors, including lisinopril. ACE inhibitors have been associated
 with a higher rate of angioedema in Black than in non-Black patients.
\end{Verbatim}
with which \fdahandler produces its answer, stating that it is done:
\begin{Verbatim}[commandchars=\\\{\}, fontsize=\small]
<DONE>
ANSWER: Yes, LISINOPRIL has been associated with an increased risk of
 angioedema. Angioedema of the face, extremities, lips, tongue,
 glottis, and/or larynx has been reported rarely in patients treated
 with angiotensin-converting enzyme inhibitors, including LISINOPRIL.
  This can occur at any time during treatment. The incidence of
 angioedema is higher in Black patients compared to non-Black
 patients. In very rare cases, fatalities have been reported due to
 angioedema associated with laryngeal edema or tongue edema. Patients
 with a history of angioedema unrelated to ACE-inhibitor therapy may
 be at increased risk of angioedema while receiving an ACE inhibitor.
  Intestinal angioedema has also been reported in patients treated
 with ACE inhibitors, presenting with abdominal pain, with or without
 nausea or vomiting.

SOURCE: LISINOPRIL label
EXTRACT_START_END: Head and Neck Angioedema: Angioedema of the
 face...associated with laryngeal edema or tongue edema.
\end{Verbatim}

\interactionagent expresses its reasoning in structured form, which is sent to the Critic:
\begin{Verbatim}[commandchars=\\\{\}, fontsize=\small]
FUNC: \{
  "name": "final_answer",
  "to": "",
  "arguments": \{
    "question": "Does LISINOPRIL increase or decrease the risk of angioedema?",
    "steps": [
      "I asked FDAHandler if LISINOPRIL increases the risk of angioedema.",
      "FDAHandler confirmed that LISINOPRIL has been associated with
      an increased risk of angioedema. Angioedema involving the face,
      extremities, lips, tongue, glottis, and/or larynx has been
      reported rarely in patients treated with angiotensin-converting
      enzyme inhibitors, including LISINOPRIL. The incidence is higher
      in Black patients compared to non-Black patients, and very rare
      cases have resulted in fatalities. Patients with a history of
      angioedema unrelated to ACE-inhibitor therapy may be at
      increased risk of angioedema while receiving an ACE
      inhibitor. Intestinal angioedema has also been reported rarely."
    ],
    "answer": "LISINOPRIL increases the risk of angioedema."
  \}
\}
\end{Verbatim}

This is processed into natural language form before being forwarded to the LLM.
\begin{Verbatim}[commandchars=\\\{\}, fontsize=\small]
        Question: Does LISINOPRIL increase or decrease the risk of angioedema?
        -----
        Reasoning:
1. I asked FDAHandler if LISINOPRIL increases the risk of angioedema.
2. FDAHandler confirmed that LISINOPRIL has been associated with an
 increased risk of angioedema. Angioedema involving the face,
 extremities, lips, tongue, glottis, and/or larynx has been reported
 rarely in patients treated with angiotensin-converting enzyme
 inhibitors, including LISINOPRIL. The incidence is higher in Black
 patients compared to non-Black patients, and very rare cases have
 resulted in fatalities. Patients with a history of angioedema
 unrelated to ACE-inhibitor therapy may be at increased risk of
 angioedema while receiving an ACE inhibitor. 
 Intestinal angioedema has also been reported rarely.
        -----
        Final answer: LISINOPRIL increases the risk of angioedema.
\end{Verbatim}

The Critic has no feedback (which indicates acceptance):
\begin{Verbatim}[commandchars=\\\{\}, fontsize=\small]
FUNC: \{
  "name": "feedback",
  "to": "",
  "arguments": \{
    "critique": ""
 \}
\}
\end{Verbatim}

As with the structured description of the reasoning steps, this is processed into natural language:
\begin{Verbatim}[commandchars=\\\{\}, fontsize=\small]
Your reasoning is valid, no feedback was provided.
\end{Verbatim}

From here on, we will show only the natural language form for these steps. The agent then outputs its answer to the original question, with the additional context from \fdahandler.
\begin{Verbatim}[commandchars=\\\{\}, fontsize=\small]
<DONE>

LISINOPRIL increases the risk of angioedema. This has been confirmed
 by FDAHandler, who referenced the LISINOPRIL label indicating that
 angioedema of the face, extremities, lips, tongue, glottis, and/or
 larynx has been reported rarely in patients treated with
 angiotensin-converting enzyme inhibitors, including LISINOPRIL. The
 incidence of angioedema is higher in Black patients compared to
 non-Black patients, and in very rare cases, fatalities have been
 reported due to angioedema associated with laryngeal edema or tongue
 edema. Patients with a history of angioedema unrelated to
 ACE-inhibitor therapy may be at increased risk of angioedema while
 receiving an ACE inhibitor. Intestinal angioedema has also been
 reported rarely in patients treated with ACE inhibitors.
\end{Verbatim}

\subsubsection{\labelagent Step}

Together with the query ``Does the angiotensin converting enzyme inhibitor category of drugs increase the risk of angioedema, decrease it, or is there no clear effect?'' the 
agent is provided with the information from the outputs from the last step, for each of the three drugs, as follows:

\begin{Verbatim}[commandchars=\\\{\}, fontsize=\small]
                Passages:
                1. Drug LISINOPRIL: LISINOPRIL increases the risk of
 angioedema. This has been confirmed by FDAHandler, who referenced the
 LISINOPRIL label indicating that angioedema of the face, extremities,
 lips, tongue, glottis, and/or larynx has been reported rarely in
 patients treated with angiotensin-converting enzyme inhibitors,
 including LISINOPRIL. The incidence of angioedema is higher in Black
 patients compared to non-Black patients, and in very rare cases,
 fatalities have been reported due to angioedema associated with
 laryngeal edema or tongue edema. Patients with a history of
 angioedema unrelated to ACE-inhibitor therapy may be at increased
 risk of angioedema while receiving an ACE inhibitor. Intestinal
 angioedema has also been reported rarely in patients treated with ACE
 inhibitors.
2. Drug CAPTOPRIL: CAPTOPRIL increases the risk of angioedema. This
 has been reported in approximately one in 1000 patients, and can
 involve various parts of the body including extremities, face, lips,
 mucous membranes, tongue, glottis, or larynx. Fatal airway
 obstruction has occurred in some cases, and the risk is higher in
 black patients compared to non-black patients.
3. Drug ENALAPRIL MALEATE: The final answer to the question "Does
 ENALAPRIL MALEATE increase or decrease the risk of angioedema?" is
 that ENALAPRIL MALEATE increases the risk of angioedema. This
 information is based on the response from FDAHandler, which
 referenced the ENALAPRIL MALEATE label indicating that angioedema has
 been reported in patients receiving the drug. The incidence is higher
 in black patients compared to non-black patients, and angioedema
 associated with laryngeal edema can be fatal. If angioedema occurs,
 treatment with ENALAPRIL MALEATE should be discontinued and
 appropriate therapy instituted immediately.
                ---------
                Does the angiotensin converting enzyme inhibitor
 category of drugs increase the risk of angioedema, decrease it, or is
 there no clear effect?
\end{Verbatim}

\labelagent immediately concludes that ACE inhibitors increase the risk of angioedema,  and sends its reasoning to the Critic:

\begin{Verbatim}[commandchars=\\\{\}, fontsize=\small]
        Question: Does the angiotensin converting enzyme inhibitor
 category of drugs increase the risk of angioedema, decrease it, or is
 there no clear effect?
        -----
        Reasoning:
        1. All three passages provided indicate that the respective
 angiotensin converting enzyme inhibitors (LISINOPRIL, CAPTOPRIL, and
 ENALAPRIL MALEATE) increase the risk of angioedema.
2. Passage 1 states that LISINOPRIL increases the risk of angioedema,
 with rare reports of angioedema in patients treated with ACE
 inhibitors, including LISINOPRIL.
3. Passage 2 indicates that CAPTOPRIL increases the risk of
 angioedema, with an incidence reported in approximately one in 1000
 patients.
4. Passage 3 confirms that ENALAPRIL MALEATE increases the risk of
 angioedema, as reported in patients receiving the drug.
5. The increased risk of angioedema is consistent across all three
 drugs mentioned, which are part of the angiotensin converting enzyme
 inhibitor category.
6. The evidence is strong as it is based on drug labels and FDAHandler
 references, which are authoritative sources.
        -----
        Final answer: The angiotensin converting enzyme inhibitor
 category of drugs increases the risk of angioedema.
\end{Verbatim}

The critic approves:
\begin{Verbatim}[commandchars=\\\{\}, fontsize=\small]
        Feedback: The reasoning provided is sound and consistent with
 medical knowledge. Angiotensin converting enzyme inhibitors are known
 to increase the risk of angioedema. The user has correctly identified
 that all three passages from the FDA data indicate an increased risk
 of angioedema with the use of LISINOPRIL, CAPTOPRIL, and ENALAPRIL
 MALEATE, which are all ACE inhibitors. The conclusion that the ACE
 inhibitor category of drugs increases the risk of angioedema is
 supported by the evidence provided and is in line with established
 medical understanding.

        If any flaws in the reasoning used to produce your answer were
 identified, you must try again.
\end{Verbatim}
       
And the agent outputs ``increase'' with very high confidence:
\begin{Verbatim}[commandchars=\\\{\}, fontsize=\small]
FUNC: \{
  "name": "category_effect_tool",
  "to": "",
  "arguments": \{
    "label": "increase",
    "confidence": 1,
    "probability": 0.001,
    "frequency": "rare",
    "evidence": "strong",
    "justification": "The evidence from FDAHandler and drug labels for
 LISINOPRIL, CAPTOPRIL, and ENALAPRIL MALEATE consistently reports an
 increased risk of angioedema with the use of these ACE
 inhibitors. The incidence of angioedema is reported as rare, with
 occurrences such as one in 1000 patients for CAPTOPRIL. The evidence
 is considered strong due to the authoritative nature of the sources."
  \}
\}
\end{Verbatim}

\subsubsection{\labelagent and Critic}
\label{app:agent-critic}
Here we present the full prompt of interactions between \labelagent and Critic, given the task of identifying the association between ``benzodiazepine'' and ``hip fracture''.

At round 1, the output from \labelagent:
\begin{lstlisting}
Question: Does the benzodiazepine category of drugs increase the risk of hip fracture, decrease it, 
or is there no clear effect?
-----
Reasoning:
1. The FDA label for Lorazepam does not specify information regarding the risk of hip fracture.
2. The FDA label data for Diazepam does not provide information on the risk of hip fracture.
3. The FDA label for Clonazepam mentions 'fracture traumatic' as an infrequent adverse event but 
does not establish a causal relationship with hip fractures or provide specific data on hip fracture risk.
4. Since none of the provided passages offer conclusive evidence regarding the effect of benzodiazepines 
on the risk of hip fracture, we cannot determine if the category as a whole has a clear effect 
on hip fracture risk.
-----
Final answer: There is no clear effect of the benzodiazepine category of drugs on the risk of hip fracture 
based on the provided passages.
\end{lstlisting}

Feedback by Critic:
\begin{lstlisting}
The reasoning provided is flawed. While it is correct that the FDA label for Lorazepam and Diazepam does not specify information regarding the risk of hip fracture, the mention of 'fracture traumatic' as an infrequent adverse event in the Clonazepam label should not be dismissed. Benzodiazepines are known to cause sedation and muscle relaxation, which can lead to an increased risk of falls, and consequently, hip fractures, especially in the elderly. The presence of 'fracture traumatic' as an adverse event in the Clonazepam label suggests that there may be a link between benzodiazepine use and fractures. Therefore, the conclusion should reflect a potential increase in the risk of hip fracture associated with benzodiazepines, even if the evidence is not conclusive for a causal relationship.

If any flaws in the reasoning used to produce your answer were identified, you must try again.
\end{lstlisting}

At round 2, the corrected output from \labelagent:
\begin{lstlisting}
Question: Does the benzodiazepine category of drugs increase the risk of hip fracture, decrease it,  or is there no clear effect?
-----
Reasoning:
1. The FDA label for Lorazepam does not specify information regarding the risk of hip fracture.
2. The FDA label data for Diazepam does not provide information on the risk of hip fracture.
3. The FDA label for Clonazepam mentions 'fracture traumatic' as an infrequent adverse event but 
does not establish a causal relationship with hip fractures or provide specific data on hip fracture risk.
4. Considering the pharmacological effects of benzodiazepines, which include sedation and muscle relaxation, there is a plausible mechanism by which these drugs could increase the risk of falls and consequently hip fractures, especially in the elderly.
5. The mention of 'fracture traumatic' as an infrequent adverse event in the Clonazepam label, while not conclusive, suggests that there may be a link between benzodiazepine use and fractures.
6. Given the plausible mechanism and the mention of fractures as an adverse event for Clonazepam, it is reasonable to infer that benzodiazepines may potentially increase the risk of hip fracture, even though the evidence is not definitive for a causal relationship.
-----
Final answer: The benzodiazepine category of drugs may potentially increase the risk of hip fracture, although the evidence is not definitive for a causal relationship based on the provided passages.
\end{lstlisting}

Feedback by Critic:
\begin{lstlisting}
The revised reasoning is more accurate in acknowledging the potential risk increase for hip fractures associated with benzodiazepines, given the pharmacological effects and the mention of 'fracture traumatic' as an adverse event for Clonazepam. However, the final answer should still reflect a degree of uncertainty due to the lack of conclusive evidence. It would be more appropriate to state that there is a potential increase in the risk of hip fracture with benzodiazepine use, but the evidence is not strong enough to definitively establish a causal relationship.

If any flaws in the reasoning used to produce your answer were identified, you must try again.
\end{lstlisting}

At round 3, the corrected output from \labelagent:
\begin{lstlisting}
Question: Does the benzodiazepine category of drugs increase the risk of hip fracture, decrease it, 
or is there no clear effect?
-----
Reasoning:
1. The FDA label for Lorazepam does not specify information regarding the risk of hip fracture.
2. The FDA label data for Diazepam does not provide information on the risk of hip fracture.
3. The FDA label for Clonazepam mentions 'fracture traumatic' as an infrequent adverse event but does not establish a causal relationship with hip fractures or provide specific data on hip fracture risk.
4. Considering the pharmacological effects of benzodiazepines, which include sedation and muscle relaxation, there is a plausible mechanism by which these drugs could increase the risk of falls and consequently hip fractures, especially in the elderly.
5. The mention of 'fracture traumatic' as an infrequent adverse event in the Clonazepam label, while not conclusive, suggests that there may be a link between benzodiazepine use and fractures.
6. Given the plausible mechanism and the mention of fractures as an adverse event for Clonazepam, 
it is reasonable to infer that benzodiazepines may potentially increase the risk of hip fracture, even though the evidence is not definitive for a causal relationship.
7. However, due to the lack of conclusive evidence, the final answer should reflect uncertainty regarding the causal relationship between benzodiazepine use and hip fracture risk.
-----
Final answer: There is a potential increase in the risk of hip fracture associated with benzodiazepine use, but the evidence is not strong enough to definitively establish a causal relationship based on the provided passages.
\end{lstlisting}

Feedback by Critic:
\begin{Verbatim}[commandchars=\\\{\}, fontsize=\small]
Your reasoning is valid, no feedback was provided.
\end{Verbatim}

Final output by \labelagent:
\begin{lstlisting}
"label": "increase",
"confidence": 0.6,
"probability": 0.1,
"frequency": "rare",
"evidence": "weak",
"justification": "The FDA label for Clonazepam lists 'fracture traumatic' as an infrequent adverse event, suggesting a potential link between benzodiazepine use and fractures. 
Given the pharmacological effects of benzodiazepines, which include sedation and muscle relaxation, there is a plausible mechanism by which these drugs could increase the risk of falls and consequently hip fractures, especially in the elderly. However, the evidence is not strong enough to definitively establish a causal relationship, and the FDA labels for Lorazepam and Diazepam do not provide specific information regarding hip fracture risk."
\end{lstlisting}

\subsection{Failed Behavior}

\ours, however, fails to identify the effect of alendronate on gastrointestinal ulcer hospitalization.
In particular, \interactionagent misinterprets a statement in the FDA labels indicating a lack of
evidence in clinical trials and forwards an overly broad statement that indicates that
clinical trials failed to show evidence of an effect to \labelagent; hence \labelagent outputs the
incorrect prediction of ``No Effect.''

\subsubsection{\interactionagent Step}

This step was run twice, for each of the representative drugs, Alendronate Sodium and Alendronate (noting that in this case only one drug should have been selected). We'll consider the results for Alendronate.

The original question to \interactionagent is ``Does ALENDRONATE SODIUM increase or 
decrease the risk of gastrointestinal ulcer hospitalization?''

\interactionagent queries \fdahandler:
\begin{Verbatim}[commandchars=\\\{\}, fontsize=\small]
FUNC: \{
  "name": "recipient_message",
  "to": "",
  "arguments": \{
    "intended_recipient": "FDAHandler",
    "content": "Does ALENDRONATE SODIUM increase 
    the risk of gastrointestinal ulcer hospitalization?"
  \}
\}
\end{Verbatim}
which queries the FDA label database with:
\begin{Verbatim}[commandchars=\\\{\}, fontsize=\small]
FUNC: \{
  "name": "relevant_search_extracts",
  "to": "",
  "arguments": \{
    "query": "Does ALENDRONATE increase the risk of gastrointestinal
 ulcer hospitalization?",
    "drug": "ALENDRONATE"
  \}
\}
\end{Verbatim}
which returns a number of passages from the FDA labels, including:
\begin{Verbatim}[commandchars=\\\{\}, fontsize=\small]
ALENDRONATE: warnings_and_cautions: 1 Upper Gastrointestinal Adverse
 Reactions BINOSTO, like other bisphosphonates administered orally,
 may cause local irritation of the upper gastrointestinal
 mucosa. Because of these possible irritant effects and a potential
 for worsening of the underlying disease, caution should be used when
 BINOSTO is given to patients with active upper gastrointestinal
 problems (such as known Barrett's esophagus, dysphagia, other
 esophageal diseases, gastritis, duodenitis, or ulcers). Esophageal
 adverse experiences, such as esophagitis, esophageal ulcers and
 esophageal erosions, occasionally with bleeding and rarely followed
 by esophageal stricture or perforation, have been reported in
 patients receiving treatment with oral bisphosphonates including
 alendronate sodium. In some cases these have been severe and required hospitalization.
\end{Verbatim}
and
\begin{Verbatim}[commandchars=\\\{\}, fontsize=\small]
ALENDRONATE: warnings_and_cautions: There have been post-marketing
 reports of gastric and duodenal ulcers with oral bisphosphonate use,
 some severe and with complications, although no increased risk was
 observed in controlled clinical trials.
\end{Verbatim}

Using these, \fdahandler responds with:
\begin{Verbatim}[commandchars=\\\{\}, fontsize=\small]
<DONE>
ANSWER: Alendronate, like other bisphosphonates administered orally,
 may cause local irritation of the upper gastrointestinal
 mucosa. Esophageal adverse experiences such as esophagitis,
 esophageal ulcers, and esophageal erosions, occasionally with
 bleeding and rarely followed by esophageal stricture or perforation,
 have been reported in patients receiving treatment with alendronate
 sodium. In some cases, these have been severe and required
 hospitalization. There have also been post-marketing reports of
 gastric and duodenal ulcers with oral bisphosphonate use, some severe
 and with complications, although no increased risk was observed 
 in controlled clinical trials.

SOURCE: ALENDRONATE label
EXTRACT_START_END: ...esophageal erosions, occasionally with bleeding
 and rarely followed by... ...gastric and duodenal ulcers with oral
 bisphosphonate use, some severe...
\end{Verbatim}

\interactionagent now concludes that the effect is uncertain, and sends its reasoning to the Critic.
Here is where the error occurs: the phrasing ``However, no increased risk was
observed in controlled clinical trials.'' is overly broad and suggests that alendronate was shown not to increase
the risk of \textit{all} gastrointestinal ulcers in clinical trials, while, in reality, that was shown only
for gastric and duodenal ulcers specifically. This will lead to the incorrect prediction by \labelagent.
\begin{Verbatim}[commandchars=\\\{\}, fontsize=\small]
        Question: Does ALENDRONATE increase or decrease the risk of
        gastrointestinal ulcer hospitalization?
        -----
        Reasoning:
        1. I asked FDAHandler whether ALENDRONATE increases the risk
        of gastrointestinal ulcer hospitalization.
2. FDAHandler responded that Alendronate can cause local irritation of
the upper gastrointestinal mucosa and has been associated with
esophageal adverse experiences such as esophagitis, esophageal ulcers,
and erosions, occasionally with bleeding and rarely followed by
esophageal stricture or perforation, which have required
hospitalization in some cases. There have also been post-marketing
reports of gastric and duodenal ulcers with oral bisphosphonate use,
some severe and with complications. However, no increased risk was observed 
in controlled clinical trials.
        -----
        Final answer: Alendronate may increase the risk of
        gastrointestinal ulcer hospitalization, as it has been
        associated with severe esophageal adverse experiences and
        post-marketing reports of gastric and duodenal ulcers that
        required hospitalization in some cases. However, it is
        important to note that no increased risk was observed in
        controlled clinical trials, indicating that while there is a potential risk, 
        it may not be common.
\end{Verbatim}
As the critic is unaware of the failure to correctly summarize the output from \fdahandler, it concludes that the answer is justified, returning no feedback; \interactionagent recieves:
\begin{Verbatim}[commandchars=\\\{\}, fontsize=\small]
 Your reasoning is valid, no feedback was provided.
\end{Verbatim}
and it outputs the answer with the flawed statement:
\begin{lstlisting}
<DONE>
Alendronate may increase the risk of gastrointestinal ulcer  hospitalization, as it has been associated with severe esophageal adverse experiences and post-marketing reports of gastric and duodenal ulcers that required hospitalization in some cases. 
However, it is important to note that no increased risk was observed in controlled clinical trials, indicating that while there is a potential risk, it may not be common.
\end{lstlisting}

\subsubsection{\labelagent step}

The question posed to \labelagent with the results from \interactionagent is as follows:
\begin{Verbatim}[commandchars=\\\{\}, fontsize=\small]
                Passages:
                1. Drug ALENDRONATE SODIUM: ALENDRONATE SODIUM has
 been associated with gastrointestinal adverse effects that can be
 severe and may lead to hospitalization, such as esophageal ulcers and
 erosions. However, it is important to note that while there have been
 post-marketing reports of severe gastric and duodenal ulcers with
 complications, no increased risk of gastrointestinal ulcer
 hospitalization was observed in controlled clinical
 trials. Therefore, while there is a potential risk, the degree of
 reliability of this association is not fully established, and the
 risk of hospitalization due to gastrointestinal ulcers appears to be rare.
2. Drug ALENDRONATE: Alendronate may increase the risk of
 gastrointestinal ulcer hospitalization, as it has been associated
 with severe esophageal adverse experiences and post-marketing reports
 of gastric and duodenal ulcers that required hospitalization in some
 cases. However, it is important to note that no increased risk was
 observed in controlled clinical trials, indicating that while there
 is a potential risk, it may not be common.
                ---------
                Does the alendronate category of drugs increase the
 risk of gastrointestinal ulcer hospitalization, decrease it, or is there no clear effect?
\end{Verbatim}

Given the overly broad statement from \interactionagent regarding clinical trial evidence, \labelagent immediately concludes that there is no clear effect and sends its reasoning steps to the Critic:
\begin{lstlisting}
Question: Does the alendronate category of drugs increase the risk of gastrointestinal ulcer hospitalization, decrease it, or is there no clear effect?
    -----
Reasoning:
1. Both passages mention that alendronate and alendronate sodium have been associated with gastrointestinal adverse effects,  including severe gastric and duodenal ulcers.
2. Both passages also note that these associations come from post-marketing reports, which are less controlled than clinical trials.
3. Passage 1 states that no increased risk of gastrointestinal ulcer hospitalization was observed in controlled clinical trials for alendronate sodium.
4. Passage 2 similarly notes that for alendronate, no increased risk was observed in controlled clinical trials.
5. Both passages conclude that while there is a potential risk, the risk of hospitalization due to gastrointestinal ulcers appears to be rare and may not be common.
6. The evidence suggests that while there are reports of severe cases, the controlled clinical trials did not show an increased risk, indicating that the effect is not consistent or well-established.
        -----
Final answer: There is no clear effect of the alendronate category of drugs on the risk of gastrointestinal ulcer hospitalization.
\end{lstlisting}
        
and the Critic approves:
\begin{Verbatim}[commandchars=\\\{\}, fontsize=\small]
        Feedback: The reasoning provided is consistent with the
 information from the passages. It correctly identifies that while
 there are post-marketing reports of gastrointestinal adverse effects,
 controlled clinical trials did not show an increased risk of
 gastrointestinal ulcer hospitalization for drugs in the alendronate
 category. The conclusion that there is no clear effect on the risk of
 gastrointestinal ulcer hospitalization is supported by the evidence
 provided. Therefore, there are no flaws in the reasoning process.

        If any flaws in the reasoning used to produce your answer were identified, 
        you must try again.
\end{Verbatim}

Finally, \labelagent outputs the incorrect final label, with high confidence:
\begin{Verbatim}[commandchars=\\\{\}, fontsize=\small]
  FUNC: \{
  "name": "category_effect_tool",
  "to": "",
  "arguments": \{
    "label": "no-effect",
    "confidence": 0.8,
    "probability": 0.05,
    "frequency": "rare",
    "evidence": "weak",
    "justification": "The alendronate category of drugs has been
 associated with gastrointestinal adverse effects in post-marketing
 reports, but controlled clinical trials did not show an increased
 risk of gastrointestinal ulcer hospitalization. The evidence suggests
 a potential risk, but it is not consistent or well-established, and
 the risk of hospitalization due to gastrointestinal ulcers appears to be rare 
 and may not be common."
  \}
\}
\end{Verbatim}

\section{{Ablations}}
\label{app:ablation}

The objective of our ablation study is to examine the individual impact of two principal elements on our system's performance:
\begin{adjustwidth}{1cm}{}
\begin{enumerate}[label=({{\arabic*}})]
    \item the iterative refinement process among multiple agents (\ie Agent-Critic interaction) 
    \item the incorporation of external knowledge via RAG.
\end{enumerate}
\end{adjustwidth}
We do not conduct an ablation of our multi-agent design as MALADE cannot be naturally expressed as a single-agent system; while expressing all agents and control flow of MALADE in a single highly complex agent is possible in principle, it is not straightforward to achieve that with any level of reliability.

For (1), we evaluate the system's performance with and without the involvement of the Critics paired with \interactionagent and \labelagent, respectively, by toggling them on and off individually. Refer to the first two columns in Table~\ref{tab:ablation}. 
For compuational reasons we do not ablate the critic on \drugagent.

For (2), we substitute FDAHandler with a simple agent which answers the questions from \interactionagent purely based on LLM's internal knowledge and generates responses in a similar output format as FDAHandler.m
Refer to the third column, labeled as ``RAG'', in Table~\ref{tab:ablation}.

Results are obtained by the \omoptask evaluation with the corresponding modified versions of MALADE, all of which were run with GPT-4 Turbo. 
To alleviate the computational burden of ablations, when an ablated system's configuration is identical to MALADE's (\ie Critics on all agents and RAG enabled) up to a given step of the pipeline, we retain the output originally produced by MALADE. 
We address the effects of variance due to random sampling from the LLM in Appendix~\ref{app:variance}.
We maintain consistency with the evaluation metrics and output label post-processing as detailed in Section~\ref{sec:omop-setup}, reporting ADE and effect-based AUC scores (with both the output confidence scores and probabilities) and ADE- and effect-based F1 scores.

\begin{table}[thb]
\centering
\begin{adjustbox}{width=\linewidth}
\begin{tabular}{cc | c | lll | lll | ll}
\toprule
\multicolumn{2}{c|}{Critics} & \multirow{2}{*}{RAG} & \multicolumn{3}{c|}{ADE-based AUC} & 
\multicolumn{3}{c|}{Effect-based AUC} & \multicolumn{2}{c}{F1 Score}\\
\cline{1-2}\cline{4-11}
\interactionagent & \labelagent & & Confidence & Probability & Probability (Modified) &
Confidence & Probability & Probability (Modified)
& ADE & Effect\\
\midrule
\checkmark&\checkmark&\checkmark&0.8514&0.7935&0.8043&0.8306&0.8058&0.8151&0.5556&0.6087\\
\checkmark&$\times$&\checkmark&0.8647&0.7126&0.7693&0.8512&0.7376&0.7862&0.5714&0.6154\\
$\times$&\checkmark&\checkmark&0.9034&0.8889&0.9143&0.8864&0.8833&0.9050&0.6316&0.6667\\
$\times$&$\times$&\checkmark&0.8249&0.8865&0.8804&0.8192&0.8812&0.8760&0.5556&0.6087\\
\checkmark&\checkmark&$\times$&0.9239&0.7609&0.8659&0.9287&0.7955&0.8853&0.5263&0.6087\\
\checkmark&$\times$&$\times$&0.9239&0.7633&0.8599&0.9287&0.7975&0.8802&0.5556&0.6364\\
$\times$&\checkmark&$\times$&0.9203&0.7403&0.8623&0.9256&0.7779&0.8822&0.5263&0.6087\\
$\times$&$\times$&$\times$&0.9203&0.7428&0.8563&0.9256&0.7800&0.8771&0.5556&0.6364\\
\bottomrule
\end{tabular}
\end{adjustbox}
\vspace{.2in}
\caption{Ablation results on MALADE.}
\label{tab:ablation}
\end{table}

Results in Table~\ref{tab:ablation} show that, in the case with RAG, the best results are obtained with the Critic on \labelagent active but the Critic on \interactionagent disabled (row 3); this suggests that the feedback from \labelagent's Critic is important for producing the most reliable results, but that \interactionagent's Critic may reduce the performance of \ours. 
The case with a Critic on \interactionagent but not \labelagent does not confirm this hypothesis (row 2), however, showing slightly improved confidence scores and F1 scores, but much worse results in terms of probabilities compared to full \ours (row 1). 
Compared to this case, removing the \interactionagent's Critic worsens results with the exception of probability-based evaluations; hence, it is difficult to confidently determine from these results whether \interactionagent's Critic is helpful or harmful. The extremely strong results in the case with the Critic on \labelagent active but the Critic on \interactionagent disabled do, however, appear to outweigh the improvements observed in the second row, suggesting that the Critic on \labelagent does in fact improve the overall reliability of MALADE.

The results without RAG show, slightly improved AUCs in the presence of \interactionagent's Critic; \labelagent's Critic, on the other hand, reduces F1 scores and slightly reduces probability-based AUCs.

Despite strong performance observed with probability-based metrics with some settings, these results suggest that direct estimates of effect probabilities may not be reliable measures in future pharmacovigilance systems; to see this, compare the columns labeled ``Probability'' and ``Probability (Modified).'' The ``Probability (Modified)'' column is the same as ``Probability'' except that output probabilities are incremented by 1 when the label is ``increase'' for ADE-based AUC and ``increase'' or ``decrease'' for effect-based AUC. In the ADE case, this modification enforces the separation between the derived scores from samples GPT-4 Turbo labeled as increasing risk and as having no effect; the improved results observed indicate that GPT-4's probability estimates are not consistent: substantial numbers of ``no-effect'' cells are assigned higher probabilities of an effect occurring as compared to cells where GPT-4 \textit{itself} identifies increased risk.

\section{{Variance of MALADE's outputs}}
\label{app:variance}

We wish to evaluate how much random sampling from LLM outputs affects \ours's outputs, and, in particular, given the potential unreliability of numerical outputs produced by LLMs~\cite{xiong2024can}, how much variance there is in \ours's output confidence scores. 
Moreover, we aim to understand whether key components of \ours's design, namely Critic agents and RAG, affect these numerical outputs, and, in particular, affect their consistency, as well as whether variance in the outputs by the first two agents in MALADE's pipeline (\drugagent and \interactionagent) is a significant contributor to the overall variance of these outputs.

\begin{figure}[t]
  \centering
  \includegraphics[width=0.8\textwidth]{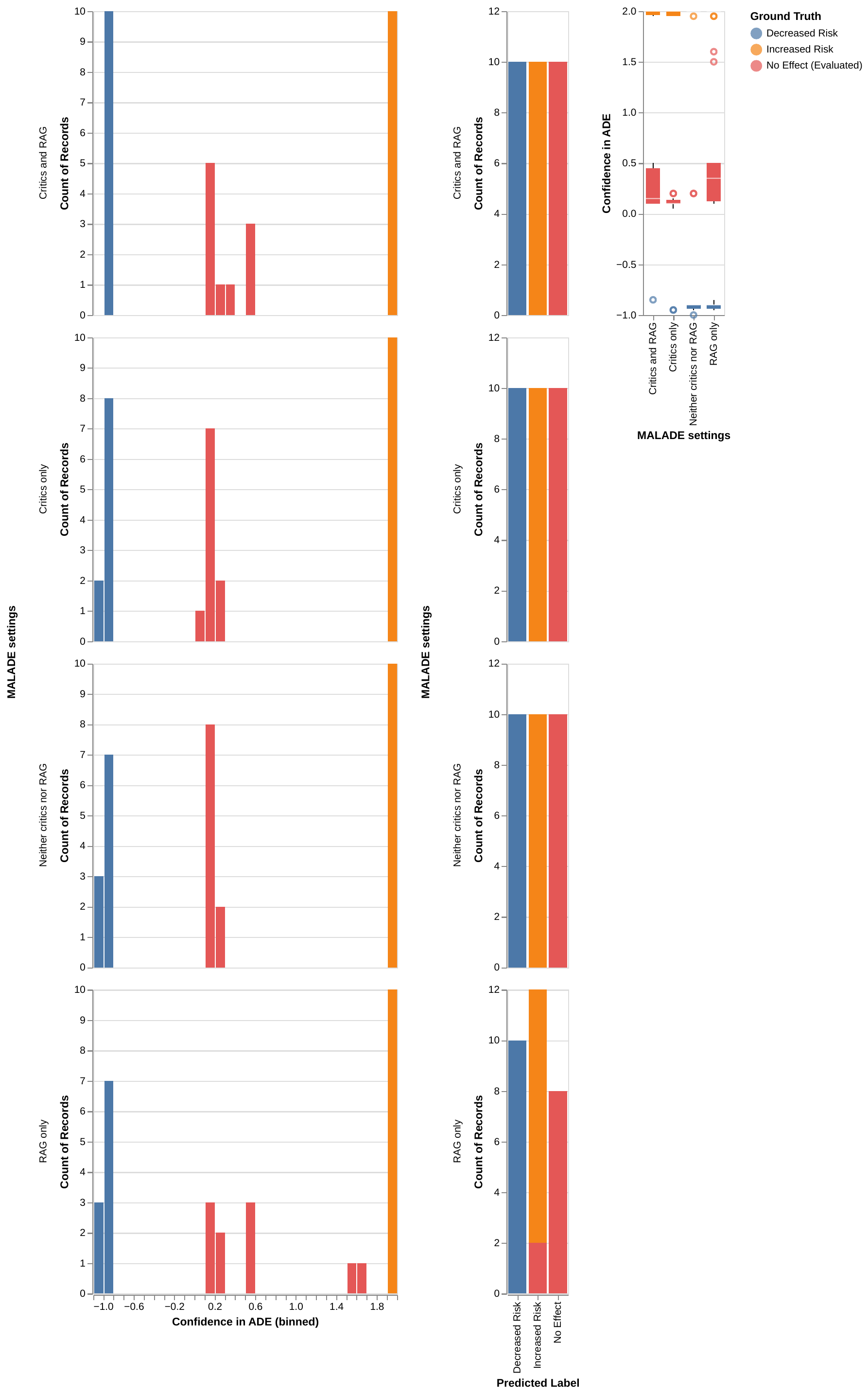}
  \caption{Histograms of confidence in ADE produced by ablations of \ours.}
  \label{fig:omop_variance_ablated}
\end{figure}

We proceed by selecting three representative cells from the OMOP table, restricting ourselves to the cells used for evaluation and, to ensure a well-defined ground truth label for each representative, to drug categories without subcategories. 
Each representative corresponds to one of the three ground truth labels (increased risk, decreased risk, and no effect). We then run ten trials on each cell with ablated versions of MALADE (constructed as in Appendix~\ref{app:ablation}; however, we only consider enabling or disabling Critics on all agents, including \drugagent here). 
The results are shown in Figure~\ref{fig:omop_variance_ablated}; these experiments were run with GPT-4 Turbo.

We observe that in all cases, MALADE maintains a clear separation between the confidences for each ground truth label, with the sole exception being the case with RAG but without Critics; that case is the only one in which we observe any samples with incorrect labels; the variance is similarly increased significantly in that case.

\begin{figure}[t]
  \centering
  \includegraphics[width=0.8\textwidth]{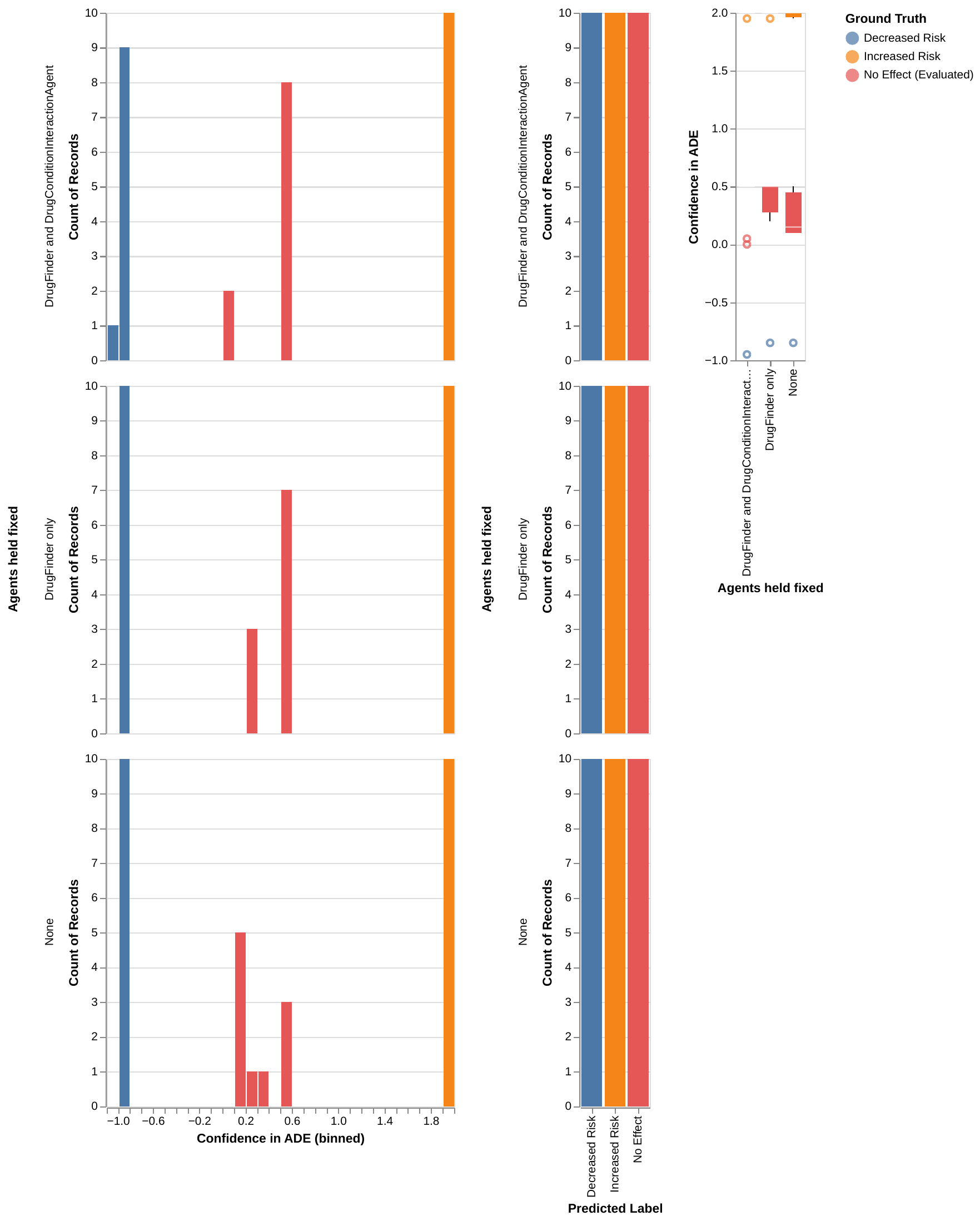}
  \caption{Histograms of confidence in ADE produced by \ours with the outputs of each initial sequence of agents in the pipeline held fixed.}
  \label{fig:omop_variance_fix_steps}
\end{figure}

Next, we investigate how the variance in the outputs of \drugagent and \interactionagent contribute to the overall variance of MALADE. 
We compare the variance of \ours's outputs with the initial steps of the pipeline held fixed;
They indicate that holding the initial steps of the pipeline fixed does not substantially reduce variance and that \labelagent is the primary source of variance in \ours. 
However, note that the variance for ``no-effect'' is, somewhat surprisingly, highest with the outputs of \drugagent and \interactionagent held constant. We observe that the variance in the representative drugs affects output confidence, in particular, it affects mean confidence in ADEs for the ``no effect'' representative. With the representatives held fixed, that mean confidence is higher (or, equivalently, mean confidence in ``no effect'' is lower) compared to the case in which we resample representatives in each trial.

\begin{table}[thb]
\centering
\begin{tabular}{c | c |  c | c}
\toprule
&Critics and RAG&Critics only&RAG only\\
\hline
Critics only&$<$, $p=0.044$&---&---\\
\hline
RAG only&$<$, $p=0.098$&$\neq$, $p=1.000$&---\\
\hline
Neither critics nor RAG&$<$, $p=0.027$&$<$, $p=0.224$&$<$, $p=0.255$\\
\bottomrule
\end{tabular}
\caption{Relationship of mean confidence in ADE for ``decrease'' for ablated versions of MALADE, with p-values.}
\label{tab:variance-decrease}
\end{table}

\begin{table}[thb]
\centering
\begin{tabular}{c | c |  c | c}
\toprule
&Critics and RAG&Critics only&RAG only\\
\hline
Critics only&$<$, $p=0.022$&---&---\\
\hline
RAG only&$>$, $p=0.076$&$>$, $p=0.017$&---\\
\hline
Neither critics nor RAG&$<$, $p=0.022$&$\neq$, $p=1.000$&$<$, $p=0.017$\\
\bottomrule
\end{tabular}
\caption{Relationship of mean confidence in ADE for ``no-effect'' for ablated versions of MALADE, with p-values.}
\label{tab:variance-no-effect}
\end{table}

\begin{table}[thb]
\centering
\begin{tabular}{c | c |  c | c}
\toprule
&Critics and RAG&Critics only&RAG only\\
\hline
Critics only&$<$, $p=0.330$&---&---\\
\hline
RAG only&$>$, $p=0.314$&$>$, $p=0.178$&---\\
\hline
Neither critics nor RAG&$>$, $p=0.144$&$>$, $p=0.067$&$>$, $p=0.278$\\
\bottomrule
\end{tabular}
\caption{Relationship of mean confidence in ADE for ``increase'' for ablated versions of MALADE, with p-values.}
\label{tab:variance-increase}
\end{table}

Now, to understand the significance of these effects, we will perform paired t-tests for each pair of ablated variants of MALADE, for each representative. The results for ``decrease'' are in Table~\ref{tab:variance-decrease}, results for ``no-effect'' are shown in Table~\ref{tab:variance-no-effect}, and results for ``increase'' are shown in Table~\ref{tab:variance-increase}. Overall, we have that the mean confidence in ADE for the representative for ``decrease'' is lowest (i.e. the confidence in ``decrease'' is highest) in the case with neither Critics nor RAG, and we have that confidence in ADE for the representative of ``no-effect'' is lower (and so confidence in ``no-effect'' is higher) in the case that we have neither Critics nor RAG as compared to the cases with both Critics and RAG and RAG alone and, in addition, that confidence in ADE is \textit{increased} (with RAG alone as compared with Critics alone, all with p-values below 0.05 for each pair.

With p-values below 0.1, we additionally have that mean confidence in ADE for ``decrease'' is decreased in the case with RAG alone as compared to the case with both Critics and RAG, confidence in ADE is increased with RAG alone as compared to critics and RAG, and, finally, that confidence in ADE for ``increase'' is increased with neither Critics nor RAG as compared to Critics alone.

Note that, while, as seen in Figure~\ref{fig:omop_variance_ablated}, by far the largest absolute shift in confidence occurs between RAG alone and all others for the ``no-effect'' representative, the large variance observed in that case is responsible for the reduced significance.

Extrapolating from these representative samples, these results suggest that \ours without RAG or Critics performs at least as well as any other configuration (with a p-value $< 0.05$), but that in the presence of RAG, Critics improve reliability on instances with no strong effect, i.e. those where the ground truth is ``no-effect,'' with a p-value less than 0.1. Notably, we do not have clear evidence that Critics are necessary when there is clear evidence, unsurprising as the FDA labels may explicitly state that a condition $H$ is a potential ADE of a drug category $C$ or that drugs in $C$ are indicated for $H$.

As discussed in Section~\ref{sec:rq2}, we consider RAG an essential component of a generalizable pharmacovigilance system. Hence, we focus on the results in the case with RAG, in which case these results suggest that the Critic components of \ours improve reliability.

\end{document}